\pdfoutput=1

\documentclass[11pt]{article}

\usepackage{acl}
\usepackage{times}
\usepackage{latexsym}
\usepackage{makecell}
\usepackage{bm}
\usepackage[T1]{fontenc}

\usepackage[utf8]{inputenc}
\usepackage{subfigure}
\usepackage{microtype}
\usepackage{enumitem}
\usepackage{graphicx}
\usepackage{multirow}
\usepackage{amsmath}
\usepackage{amssymb}
\usepackage{algorithm}
\usepackage{algorithmicx}
\usepackage{algpseudocode}
\usepackage{array}
\usepackage{booktabs} 
\usepackage{arydshln}
\usepackage{amsmath}
\usepackage{amssymb}
\usepackage{mathtools}
\usepackage{amsthm}
\usepackage{arydshln}
\usepackage{url}
\usepackage{listings}
\title{LLM-Powered Benchmark Factory: Reliable, Generic, and Efficient}
\author{Peiwen Yuan$^1$, Shaoxiong Feng$^2$, Yiwei Li$^1$, Xinglin Wang$^1$, Yueqi Zhang$^1$\\ {\bf Jiayi Shi$^1$, Chuyi Tan$^1$, Boyuan Pan$^2$, Yao Hu$^2$, Kan Li$^{1}$\footnotemark[1]}\\
  $^1$School of Computer Science and Technology, Beijing Institute of Technology \\
  $^2$Xiaohongshu Inc \\
  \texttt{\{peiwenyuan,liyiwei,wangxinglin,zhangyq,shijiayi,tanchuyi\}@bit.edu.cn} \\ 
  \texttt{\{likan\}@bit.edu.cn} \ \ 
  \texttt{\{shaoxiongfeng2023\}@gmail.com} \\  \texttt{\{panboyuan,xiahou\}@xiaohongshu.com}}

\begin{document}
\maketitle
\renewcommand{\thefootnote}{\fnsymbol{footnote}} 
\footnotetext[1]{Corresponding author.} 
\renewcommand{\thefootnote}{\arabic{footnote}}
\begin{abstract}
The rapid advancement of large language models (LLMs) has led to a surge in both model supply and application demands. To facilitate effective matching between them, reliable, generic and efficient benchmark generators are widely needed.
However, human annotators are constrained by inefficiency, and current LLM benchmark generators not only lack generalizability but also struggle with limited reliability, as they lack a comprehensive evaluation framework for validation and optimization.
To fill this gap, 
we first propose an automated and unbiased evaluation framework, structured around four dimensions and ten criteria. Under this framework, we carefully analyze the advantages and weaknesses of directly prompting LLMs as generic benchmark generators. To enhance the reliability, we introduce a series of methods to address the identified weaknesses and integrate them as \textsc{BenchMaker}. 
Experiments across multiple LLMs and tasks confirm that \textsc{BenchMaker} achieves superior or comparable performance to human-annotated benchmarks on all metrics, highlighting its generalizability and reliability. More importantly, it delivers highly consistent evaluation results across 12 LLMs (0.967 Pearson correlation against MMLU-Pro), while taking only \$0.005 and 0.38 minutes per sample. See our codes in \url{https://github.com/ypw0102/BenchMaker}.
\end{abstract}
\section{Introduction}
\label{sec:intro}

\begin{figure}[t]
\centering
\includegraphics[width=0.48\textwidth]{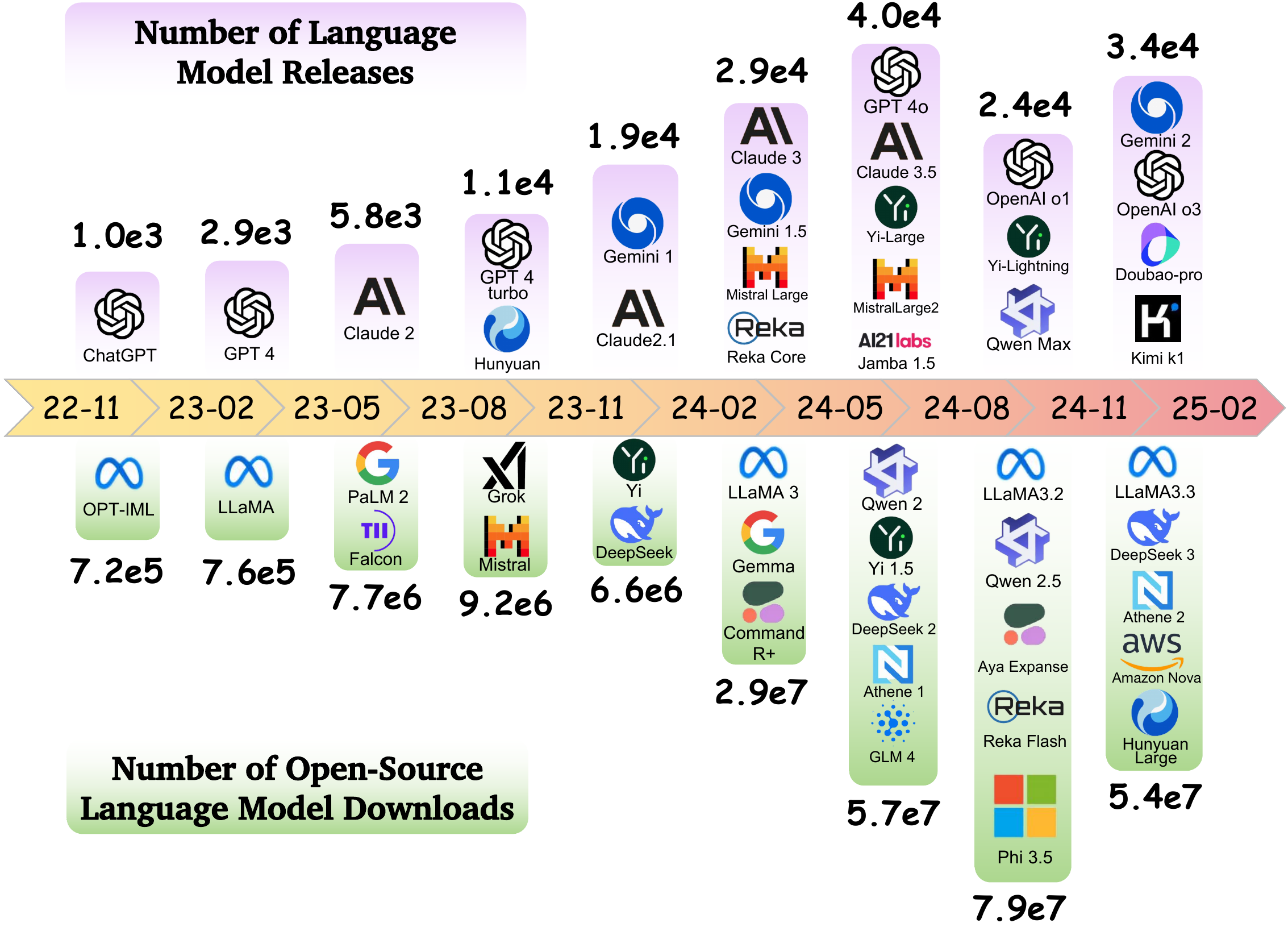}
\caption{The trends of LLMs released and open-source LLMs downloads per season since the debut of ChatGPT. We obtain the data via the Huggingface API. See details in Appendix~\ref{sec:huggingface}.}
\vspace{-11pt}
\label{fig:abs}
\end{figure}

With the ongoing scaling up of large language models (LLMs) in multiple dimensions over the past few years, two key trends have emerged (Figure~\ref{fig:abs}): 
(1) The LLM release process has accelerated and now exceeds 30k per season; (2) The growth in LLM capabilities has spurred application demand, reflected in over 50M downloads of open-source models per season.
Serving as a bridge between massive LLM supply and various application needs, the demand for customized benchmarks is rapidly growing, helping downstream tasks identify the most suitable LLM. 

However, current benchmark construction processes largely rely on human-provided signals \citep{evalsurvey,mmlupro}, leading to long cycles and high costs. To this end, efficient LLM-driven methods have recently been explored. 
Unfortunately, they generally rely on the existence of seed benchmarks for data augmentation \citep{dyval2,unigen,perteval,databench} and task specific designs \citep{dyval,s3eval}, lacking generalization across tasks and domains. 
Meanwhile, the current absence of a comprehensive evaluation framework hinders the assessment and optimization of benchmark generators, weakening our confidence in their reliability for real applications.
Hence, an automatic and comprehensive evaluation framework and a generic and reliable benchmark generator that can handle any assessment demands and efficiently generate high-quality samples are urgently needed.

%


To this end, we first construct an automatic evaluation framework with ten criteria for benchmark generators. Notably, we utilize causal learning \citep{caulearn} techniques to identify and remove biases of LLM-as-a-judge \citep{geval} across various criteria, ensuring the reliability of the framework. 
On this basis, we examine the strengths and weaknesses of directly prompting LLM as generic benchmark generator through this evaluation framework. The results reveal that the generated benchmark exhibits limited lexical and semantic diversity, poor controllability over difficulty, and low sample faithfulness, while showing advantages in high task alignment and knowledge diversity.
Bearing this in mind, we develop a generic benchmark generator \textsc{BenchMaker} by integrating existing techniques with newly designed approaches to address the identified issues.
Specifically, \textsc{BenchMaker}: strengthens sample faithfulness using stepwise self-correction generation and conflict guided contrastive discrimination; extends difficulty boundary with difficulty strategy guidance and difficulty diffusion mechanism; enhances diversity through AttrPrompt \citep{attr} and in-batch redundancy filtering. We also discuss some unsuccessful attempts in Appendix~\ref{sec:unsuccess} to provide more insights for future research.


We conduct comprehensive experiments to validate \textsc{BenchMaker}  under the proposed evaluation framework.
Compared to high-quality human-annotated benchmarks, the benchmarks generated by \textsc{BenchMaker} exhibit superior task alignment, better difficulty controllability, more challenging question difficulties, and comparable sample faithfulness and diversity.
More importantly, they yield highly consistent evaluation results across 12 LLMs (0.967 Pearson correlation with MMLU-Pro), with \textsc{BenchMaker} taking only \$0.005 and 0.38 minutes per sample.
We further perform detailed experiments to validate the outstanding generalization and robustness across tasks and LLMs, and the effectiveness of each component of \textsc{BenchMaker}.
Finally, we derive a formula for evaluating the confidence of benchmarking results under conditions where faithfulness cannot be fully satisfied, further enhancing the practicality and reliability of \textsc{BenchMaker}.

\section{Backgrounds}
\label{sec:rel}
In this section, we first review the latest developments in data synthesis \S\ref{sec:syndata} and then discuss the potential values of developing a generic benchmark generator \S\ref{sec:app}.
\subsection{Synthetic Data Generation}
\label{sec:syndata}

The growth of language model abilities has led to widespread research on LLM-driven data synthesis, which demonstrates much better quality and controllability over traditional approaches \citep{datasurvey1,datasurvey2}. 
Centering around the construction of data flywheel (LLM-driven evolution) \citep{flywheel1,flywheel2}, training data synthesis has garnered much attention in fields like mathematics \citep{datamath}, science \citep{datasci}, and code \citep{datacode}, continuously pushing LLMs' capability boundaries.
Unlike the training data synthesis aimed at optimizing model performance, the goal of benchmark synthesis is to accurately evaluate models on specific task, presenting greater challenges in both measurement and implementation \citep{evalsurvey}. 
In terms of measurement, recent studies \citep{dyval,databench,perteval} generally focus on specific criteria, without establishing a comprehensive evaluation system for benchmark generators.
In terms of implementation, current benchmark generators \citep{modelwritten,unigen,dyval2,s3eval} are constrained by their dependence on existing benchmarks and task specific designs, preventing them from being generic. We construct a comprehensive  evaluation framework and develop generic and reliable \textsc{BenchMaker} method to fill this gap.

\subsection{Potential Applicable Scenarios of \textsc{BenchMaker}}
\label{sec:app}
Given arbitrary assessment demands $X$ as the sole input, a generic benchmark generator (\textsc{BenchMaker}) $\mathcal{G}$ is expected to generate a well-aligned high-quality benchmark $\mathcal{D}$.
On this basis, we summarize its applicable scenarios as follows: (1) Complementing existing benchmarks for tailored assessment demands; (2) Acting as a dynamic benchmark generator to alleviate data contamination issues \citep{datacontamination}; (3) Serving as a difficulty controllable benchmark generator to mitigate the benchmark saturation problem \citep{saturation}; (4) Functioning as a versatile training data generator. 
Therefore, building \textsc{BenchMaker} holds significant importance for both scientific research and practical applications within the NLP community. 
\section{Benchmarking Benchmark Generator}
\label{sec:pre}

\begin{table*}[t]
    \renewcommand\arraystretch{1.3}
    \small
    \centering
    \caption{Criteria taxonomy and definition of the proposed evaluation framework for the benchmark generator. Criteria marked with * indicate optimization objectives that are distinctive to benchmark synthesis compared to training data synthesis.}
    \setlength{\tabcolsep}{0.3em} 
    \begin{tabular}{lll}
    \toprule
    \textbf{Taxonomy}&\textbf{Criterion} & \textbf{Definition}  \\
    \midrule
    \multirow{2}{*}{Credibility}&faithfulness&The sample is well-defined, with the ground truth answer being correct.\\
    \multirow{2}{*}{}&alignment *&The abilities evaluated by the sample align well with the given assessment demands.\\
    \hline
    \multirow{3}{*}{Diversity}&lexical&The samples exhibit sufficient lexical diversity.\\
    \multirow{3}{*}{}&semantic&The samples exhibit sufficient semantic richness.\\
    \multirow{3}{*}{}&knowledge *&The knowledge and skills assessed by different samples should not be redundant.\\
    \hline
    \multirow{2}{*}{Difficulty}&controllability *&The samples have correct difficulty labels to form subsets with varying difficulties.\\
    \multirow{2}{*}{}&boundary *&The hardest subset is difficult enough to explore the boundaries of advanced models.\\
    \hline
    \multirow{3}{*}{Benchmark-Level}&effectiveness *&The benchmarking results align with human benchmark under the same assessment demands.\\
    \multirow{3}{*}{}&robustness *&The benchmarking results of generated benchmarks under similar assessment demands align.\\
    \multirow{3}{*}{}&efficiency&The time and cost of generating a benchmark are low enough.\\
    \bottomrule
    \end{tabular}
    \vspace{-3pt}
    \label{tab: criteria}
\end{table*}

\begin{figure*}[h]
\centering
\includegraphics[width=0.95\textwidth]{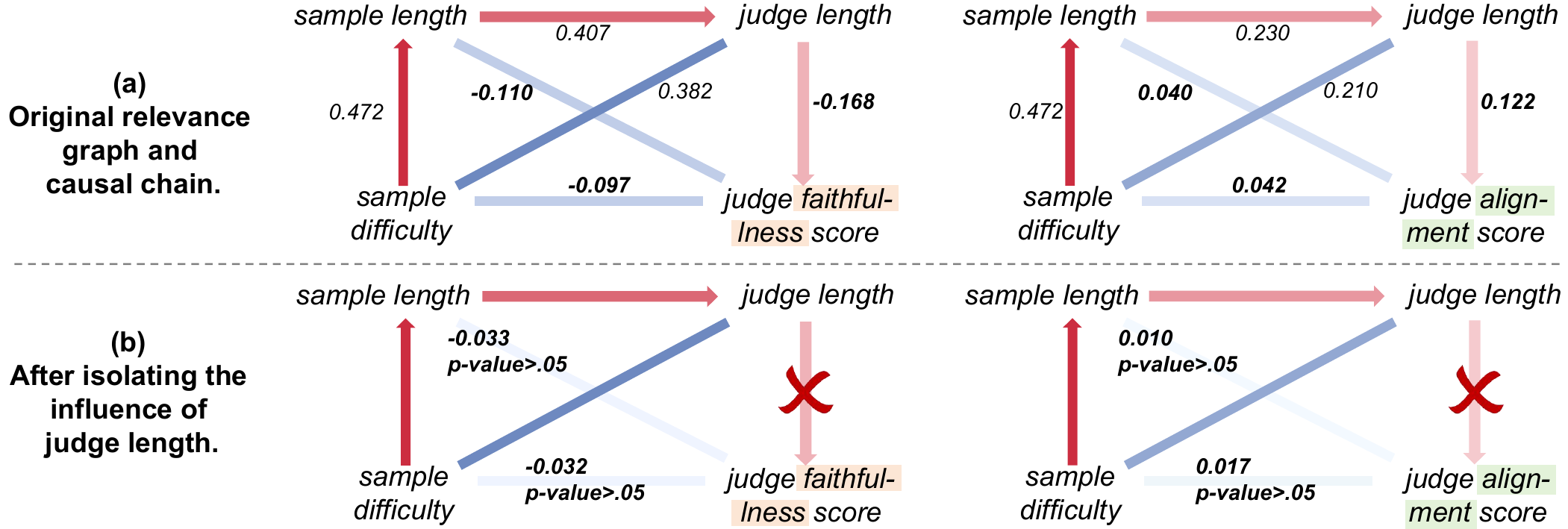}
\vspace{-10pt}
\caption{{Pearson correlations among key factors of benchmark evaluation and LLM (Qwen-Plus) judge scores (faithfulness and alignment). The most relevant path of each subject is highlighted in red to show the possible causal chain.}}
\vspace{-7pt}
\label{fig:chain}
\end{figure*}

While training data synthesis focuses on faithfulness, diversity and the final performance of the trained models \citep{attr,datasurvey2}, the evaluation of synthetic benchmark should be more comprehensive to ensure the reliability of its benchmarking results. Thus, we carefully establish an evaluation framework for benchmark generator with ten criteria, as illustrated in Table~\ref{tab: criteria}.
\subsection{Credibility}
\label{sec:credibility}
Two key criteria for ensuring the credibility of a benchmark are \textbf{faithfulness} and \textbf{alignment}. Faithfulness indicates that the generated sample be free of ambiguity with a correct answer. Alignment requires the generated samples to strictly adhere to the specified assessment demands $X$, especially in abilities to be assessed. 
For these criteria, previous approaches rely on human evaluation \citep{unigen,dyval2} or LLM-as-a-judge \citep{llmasjudge}. However, the former lacks automation, and the latter is susceptible to biases \citep{judgingjudge}. 

To this end, we seek to detect and mitigate any biases of LLM-as-a-judge that may exist within the framework. 
We choose Qwen-Plus \citep{qwen} as the judge with scoring range as $[0,1]$ (See prompt in Appendix~\ref{sec:prompt_judge}). Experiments are conducted on the high-quality MATH benchmark \citep{math}, for which we assign score 1 to both faithfulness and alignment for every sample. 
Ideally, the scores assigned by the judge should not exhibit any consistency with specific factors. However, as shown in Figure~\ref{fig:chain}-(a), both faithfulness and alignment are significantly correlated (p-value $<$ 0.05) with sample difficulty, sample length, and the length of the judge’s rationale. 
For each factor, we highlight its weightiest path in red, revealing a clear causal chain: harder questions lead to longer samples, requiring judges to conduct lengthier analyses. For faithfulness, longer analyses increases the likelihood of judge errors, resulting in lower faithfulness ratings. While for alignment, longer analyses increases the probability of task-relevant words appearing and results in higher alignment ratings.
To validate the above hypothesis, we control the judge length and respectively calculate the partial correlations \citep{pengouin} of sample difficulty and sample length with faithfulness and alignment. 
As shown in Figure~\ref{fig:chain}-(b), after isolating the influence of judge length, the effects of other factors are no longer significant (p-value $>$ 0.05). Similar conclusions also hold true when GPT-4o mini \citep{4o} serves as the judge (Figure~\ref{fig:chain_4omini} in Appendix).

Based on the analysis above, the potential biases of the LLM judge in this scenario are all mediated by judgment length.
Therefore, for benchmark generators $\mathcal{G}_{1:|\mathcal{G}|}$ under evaluation, we derive their unbiased judge results with a Multiple Regression model. Specifically, we set the judge score as the dependent variable, the generator categories as dummy variables, and judge length as the covariate:
\begin{equation}
\small
f(i) = \beta_i + \beta_{len} \cdot \text{judge\_length}+\epsilon
\label{equation: bias}
\end{equation}
where $f(i)$ denotes the average judge score of $\mathcal{G}_{i}$ and $\beta_i$ reflects the debiased score of $\mathcal{G}_{i}$, which we select as our metrics for faithfulness and alignment.

\subsection{Diversity}
\label{sec:diversity}
With credibility ensured, the diversity of the benchmark determines the extent to which evaluation results can reflect the true model capability across the assessed domain. 
Apart from the widely tested lexical and semantic diversity, our framework also examines the knowledge diversity to make the evaluation more comprehensive.

\paragraph{Lexical Diversity}reflects vocabulary richness in benchmarks. Traditional metrics like vocabulary size and self-BLEU \citep{selfbleu} used in \citet{unigen} and \citep{attr} are biased by sample length \citep{diverbias}. We use unbiased word frequency entropy \citep{entropy} as the metric to evaluate lexical diversity.


\paragraph{Semantic Diversity}quantifies a benchmark's semantic comprehensiveness. We calculate the average Euclidean distance between semantic embeddings of samples as the metric. Specifically, we use powerful text-embedding-ada-002 \citep{text-embedding-ada-002} as the embedding model.

\paragraph{Knowledge Diversity}evaluates whether the samples evaluate different sub-abilities within the assessment demands. 
When samples test the same sub-ability, the model is likely to exhibit similar correctness patterns. Therefore, we use the correctness of a set of models (denoted as $\mathcal{M}_{1:|\mathcal{M}|}$, see Appendix~\ref{sec:bench_model_list} for detailed list) to represent the knowledge embedding for each sample. If the embeddings of two samples are highly similar, it reflects a strong alignment in the sub-abilities they assess. The average pairwise Hamming distance \citep{hamming}, suitable for discrete embeddings, is employed as the metric for this criterion.

\subsection{Difficulty}
\label{sec:difficulty}
When diversity meets requirements, we should further consider the difficulty attribute, which is particularly significant in an era of increasingly divergent model capabilities.
\paragraph{Difficulty Controllability}refers to assigning differentiated difficulty labels to the samples (e.g., MATH \citep{math}). These labels enable the benchmark to be divided into subsets for more targeted evaluation of models with varying capabilities. For each sample, we use the average error rate of $\mathcal{M}_{1:|\mathcal{M}|}$ as the ground truth for difficulty label. Based on this, we compute the Spearman correlation between the difficulty labels provided by the benchmark and the ground truth as the metric.
\paragraph{Difficulty Boundary} denotes the difficulty of the hardest subset of a benchmark.
With the growing strength of LLMs, their performance on simpler benchmarks has reached saturation \cite{mmlu}, making it difficult to differentiate their capabilities.
Consequently, more challenging benchmarks \citep{mmlupro} are continuously introduced to evaluate the latest LLMs.
Thus, we propose assessing the average error rate of $\mathcal{M}_{1:|\mathcal{M}|}$ on the hardest subset of benchmark to measure its difficulty boundary.

\subsection{Benchmark-Level}
\label{sec:benchmarklevel}
Lastly, we introduce high-level metrics for assessing benchmark generators.
\paragraph{Effectiveness.}
While the earlier criteria assess benchmark quality from various aspects, a unified metric is required to measure benchmark effectiveness.
Taking high-quality human-annotated benchmark as the ground truth, we examine whether generated benchmark under identical assessment demands can deliver equivalent evaluation results.
To this end, we calculate the accuracy of $\mathcal{M}_{1:|\mathcal{M}|}$ on both generated and human benchmarks and use the Pearson correlation between them as the effectiveness metric.

\paragraph{Robustness.}
Under similar inputs, a robust system should produce comparable outputs. Similarly, we expect a robust benchmark generator to produce benchmarks with equivalent evaluation efficacy for similar assessment demands. Thus, we calculate the accuracy of $\mathcal{M}_{1:|\mathcal{M}|}$ on benchmarks generated under similar assessment demands (the original and that rewritten by GPT-4o) and calculate the Pearson correlation between them as the robustness metric.

\paragraph{Efficiency.} 
High-quality human-annotated benchmarks are constrained due to inefficiencies in their construction. We evaluate the efficiency of a benchmark generator by measuring the time and monetary costs associated with generating benchmarks of a certain size.

By establishing this comprehensive evaluation framework, the strengths and weaknesses of benchmark generators can be thoroughly assessed, and the reliability of the proposed method can be validated.

\section{Development of BenchMaker}
\label{submission}

\begin{figure*}[h]
\centering
\includegraphics[width=0.95\textwidth]{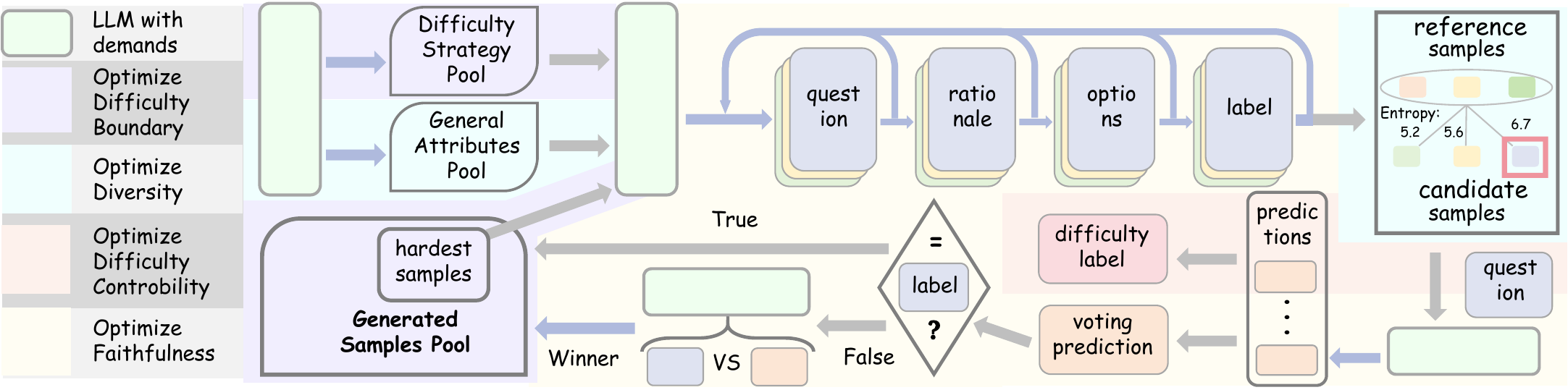}
\caption{Overview of \textsc{BenchMaker}.}
\label{fig:method}
\end{figure*}

In this section, we first discuss the primary sample format we studied in \S\ref{sec:sampleformat}.
Afterwards, since previous studies have yet to realize generic benchmark generators (with assessment demands $X$ as the sole input), we analyze the pros and cons of directly prompting the LLM as such generator in \S\ref{sec:proscons}. Building on the experimental results, we refine its weaknesses in the following sections, leading to the development of \textsc{BenchMaker}.

\subsection{Sample Format Selection}
\label{sec:sampleformat}
Following previous studies \citep{perteval,dyval2}, we have chosen multiple-choice questions (MCQs) as the primary sample format for benchmark generation based on the following reasons: (1) Versatility: MCQ serves as a universal format for evaluating most capabilities; (2) Accuracy: Misjudgment can be effectively prevented caused by variations in output formats \citep{format}; (3) Efficiency: MCQs do not depend on external modules such as LLM-as-a-Judge, ensuring a streamlined evaluation process; (4) Transformability: Each generated sample includes a rationale, enabling easy conversion into other formats, such as the text generation format presented in the Appendix~\ref{sec:convert}.

\subsection{Pros and Cons of Directly Prompting}
\label{sec:proscons}
We choose MATH \citep{math}, MMLU-Pro \citep{mmlupro} and HellaSwag \citep{hellaswag} as high-quality benchmarks $\mathcal{D}_{human}$ for comparison. We adopt the prompt in Appendix~\ref{sec:prompt_direct} to guide $\mathcal{M}$: GPT-4o mini in generating credible and diverse samples $s_{1:|\mathcal{D}_{human}|}$:
\begin{equation}
    s_i = \{q_i,r_i,o_i,a_i\} = \mathcal{M}(\text{prompt}_{base},l,X)
\end{equation}
where $q_i,r_i,o_i,a_i$ denote question, rationale, options, and label, respectively. We proportionally adjust the difficulty level $l$ from 1 to 10 in the prompt (see descriptions in Appendix~\ref{sec:dif_level}), and select samples with top 20\% difficulty level to form the hardest subset. 
The assessment demands $X$ are shown in Appendix~\ref{sec:demands}.
As shown in Table~\ref{tab:main}, compared to $\mathcal{D}_{human}$, directly prompting LLM as generic benchmark generator demonstrates poorer faithfulness, lower lexical and semantic diversity, weaker difficulty controllability, and less challenging subset. 
Meanwhile, we also observe its advantages in better alignment\footnote{We set the debiased LLM-as-a-judge score of the human benchmark to 1, adjusting scores of generated benchmarks accordingly, which may result in scores exceeding 1.}, greater knowledge diversity, and improved efficiency. 

\subsection{Faithfulness Optimization}
\label{sec:fat_opt}
To enhance faithfulness, previous studies have explored methods such as self-correction \citep{SelfIns,SelfReflect} and the use of external tools \citep{solver,rag}. As self-correction offers greater versatility, we propose the following two BenchMaker-compatible techniques to optimize faithfulness.
\paragraph{Stepwise Self-correction.}Since errors might occur at any step during the generation of $\{q_i,r_i,o_i,a_i\}$, we instruct the model to validate the content at each step. If an error is detected, the model will return to the beginning. Compared to full-sample self-checking, step-wise critique boosts error detection with less decoding cost (See Appendix~\ref{sec:unsuccess}).
\paragraph{Conflict Guided Contrastive Discrimination.} 
\citet{cannotselfcritic} finds that LLMs struggle to correctly judge their prior answers on challenging questions. 
Therefore, we extend Stepwise Self-correction by having the LLM not only act as a judge but also as a test-taker to identify potential errors. Let the LLM answers $q_i$ $T$ times to attain $\Bar{a}_i^{1:T}$, we get the self-consistency \citep{sc} result $\Dot{\Bar{a}}_i$ through majority voting. If $\Dot{\Bar{a}}_i \neq a_i$, the conflict suggests differing $r_i$ and $\Dot{\Bar{r}}_i$. As \citet{llmasjudge} finds that comparison-based judges are more accurate than item-wise judges, we have the LLM conduct a contrastive discrimination between $r_i$ and $\Dot{\Bar{r}}_i$ to determine the final rationale and label for $s_i$.

\subsection{Difficulty Optimization}
\label{sec:diff_bound_opt}
\paragraph{Difficulty Controllability.} From \S\ref{sec:proscons}, we know that the LLM's ability to control the difficulty of generated samples is limited. In particular, for the language understanding task (MMLU-Pro), the Spearman correlation between the actual and expected difficulty of the samples is only 0.021. To further explore this, we examine LLM’s difficulty perception by asking it to score the difficulty label of the generated samples. However, the correlation only increases to 0.089, suggesting that while LLM has some capacity to perceive difficulty, it is still weak. We then switch the role of LLM and assess the difficulty from the perspective of test-taker:
\begin{equation}
    \beta_i = \frac{1}{T} \sum_{j=1}^T \textbf{1}_{\Bar{a}_i^j\neq a_i}
\end{equation}
By taking the inconsistency between $\Bar{a}_i^{1:T}$ and $a_i$ as difficulty label, the correlation increases to 0.415, suggesting that $\beta$ is a reliable metric for difficulty controllability.

\paragraph{Difficulty Diffusion Mechanism.} 
Given that the LLM has a certain level of difficulty perception, we iteratively select the more challenging samples according to $\beta$ from the generated ones as difficulty references, and instruct the LLM to generate a more difficult sample. This allows the sample difficulty to rise continuously through diffusion. The detailed algorithm is described in Appendix~\ref{sec:ddd}. 
\paragraph{Difficulty Strategy Guidance.}
We further consider providing the LLM with task-specific difficulty-control strategies. Specifically, we first require the LLM to give varying strategies for generating samples of specific difficulty levels based on the given $X$ (see examples in Appendix~\ref{sec:strategies}). For example, difficult samples those assessing reasoning ability generally require more reasoning steps. With the Difficulty Diffusion Mechanism, we progressively introduce more difficult sample generation strategies to the LLM to further extend the difficulty boundary.

\subsection{Diversity Optimization}
\label{sec:div_opt}
The optimization of synthetic data diversity has been widely studied \citep{datasurvey1}. We conduct extensive tests and select the most generic and effective \textbf{AttrPrompt} \citep{attr} technique for \textsc{BenchMaker}. 
AttrPrompt explicitly enhances the lexical and semantic diversity of benchmarks by randomly assigning pre-generated (attribute, value) pairs as part of the input for each sample.
Furthermore, we notice that the introduction of treating the generated samples as difficulty references might cause sample homogeneity. To mitigate this, we propose an \textbf{In-batch Diversity Boosting} method, where LLM generates $L$ (We set $L$ as 5 for our default setting) candidate samples and selects the one with the greatest word frequency entropy difference from the input reference samples.

\section{Experiments and Analyses}
\label{sec:exp}
\begin{table*}[t]
\renewcommand\arraystretch{1.2}
\centering
\small
\caption{Overall experimental results under the proposed evaluation framework. For each setting, we run three times and report the average results. We take GPT-4o mini as default generator. Values in bold denote the best results between \textsc{BenchMaker} and Human Benchmark.}
\setlength{\tabcolsep}{0.32em} 
\begin{tabular}{ccccccccccc}
\toprule
\multirow{3}{*}{\textbf{Methods}} & \textbf{Faithful} & \textbf{Alignment} & \textbf{Lexical} & \textbf{Semantic} & \textbf{Knowledge} & \textbf{Control} & \textbf{Boundary} & \textbf{Effective} & \textbf{Robust} & \textbf{Efficiency}\\
\cdashline{2-11}
&  Unbias & Unbias  & \multirow{2}{*}{Entropy\textbf{$\uparrow$}} & Euclidean  & Hamming & Spea-  & Error&Pear- &Pear- &\$/item,\\
&Score\textbf{$\uparrow$} &Score\textbf{$\uparrow$}&&Distance\textbf{$\uparrow$}&Distance\textbf{$\uparrow$}&rman\textbf{$\uparrow$}&Rate\textbf{$\uparrow$}&son\textbf{$\uparrow$}&son\textbf{$\uparrow$}&min/item\textbf{$\downarrow$}\\
\midrule
\multicolumn{11}{c}{MATH \citep{math}} \\
Human Benchmark& \textbf{1.000} & 1.000 & 8.054 & 0.665 & 0.349 & 0.143 & 0.752&-&-&high \\
Direct Prompt& 0.665 & 1.166 & 7.091 & 0.618 & 0.365 & 0.109 & 0.635 &0.687 & 0.991 & 0.002, 0.17 \\
+AttrPrompt& 0.611 & 1.138 & 8.265 & 0.675 & 0.360 & 0.124 & 0.659 &0.759 & 0.983 & 0.002, 0.19 \\
+InBatchDivBoost& 0.623 & 1.142 & 8.652 & 0.677 & 0.366 & 0.115 & 0.628 &0.778 & 0.985 & 0.003, 0.20 \\
+StepSelfCorrect& 0.924 & 1.152 & 8.674 & 0.675 & 0.369 & 0.162 & 0.557 &0.803 & 0.979 & 0.003, 0.23 \\
+ConflictConDisc& 1.019 & 1.151 & 8.668 & 0.678 & 0.357 & 0.175 & 0.515 &0.838 & 0.992 & 0.004, 0.35 \\
+DiffControl& 1.019 & 1.151 & 8.668 & 0.678 & 0.357 & 0.403 & 0.515 &0.838 & 0.992 & 0.004, 0.35 \\
+DiffDiffusion& 0.994 & 1.166 & 8.705 & 0.680 & 0.387 & 0.451 & 0.683 &0.882 & 0.990 & 0.005, 0.39 \\
BenchMaker& 0.930 & 1.200 & \textbf{8.976} & \textbf{0.681} & 0.403 & \textbf{0.434} & 0.768&0.935&0.986&\textbf{0.005, 0.42} \\
BenchMaker$_{\text{4o}}$& 0.918 & \textbf{1.223} & 8.835 & 0.675 & 0.385 & 0.432 & \textbf{0.779}&\textbf{0.941}&\textbf{0.988}&0.084, 1.12 \\
BenchMaker$_{\text{haiku}}$& 0.902 & 1.116 & 8.878 & 0.676 & \textbf{0.410} & 0.401 & 0.775&0.912&0.979&0.026, 0.57 \\
\cdashline{1-11}
\multicolumn{11}{c}{MMLU-Pro \citep{mmlupro}} \\
Human Benchmark& 1.000 & 1.000 & \textbf{10.404} & \textbf{0.731} & 0.307 & 0.000 & 0.751&-&-&high \\
Direct Prompt& 0.894 & 1.218 & 9.608 & 0.726 &0.391  & 0.021 & 0.587&0.850&0.989&0.002, 0.16 \\
BenchMaker& \textbf{1.020} & \textbf{1.245} & 10.166 & 0.728 & \textbf{0.395} & \textbf{0.477} & \textbf{0.759} & 0.967&0.982 & \textbf{0.005, 0.38} \\
\cdashline{1-11}
\multicolumn{11}{c}{HellaSwag \citep{hellaswag}} \\
Human Benchmark& 1.000 & 1.000 & \textbf{9.167} & 0.655 & 0.384 & 0.000 & 0.569&-&-&high \\
Direct Prompt& 0.862 & 1.107 & 8.165 & 0.660 &0.396  & 0.047 & 0.626&0.821&0.979&0.002, 0.17 \\
BenchMaker& \textbf{1.032} & \textbf{1.130} & 9.052 & \textbf{0.663} & \textbf{0.421} & \textbf{0.439} & \textbf{0.708} & 0.958&0.984 & \textbf{0.005, 0.40} \\
\bottomrule
\end{tabular}
\label{tab:main}
\vspace{-0.1cm}
\end{table*}
We conduct comprehensive experiments to validate \textsc{BenchMaker} under the proposed framework in this section.
\paragraph{Settings.}
We select the widely used human-annotated MATH\footnote{Converted into a MCQ format, see details in Appendix~\ref{sec:MATH_convert}.} \citep{math} (mathematical reasoning), MMLU-Pro (multi-task language understanding) \citep{mmlupro} and HellaSwag (commonsense reasoning) \citep{hellaswag} as high-quality baseline benchmarks.
For the 7 subsets of MATH, the 13 subsets of MMLU-Pro\footnote{Excluding the type 'other'.} and HellaSwag, we write simple assessment demands respectively (see details in Appendix~\ref{sec:demands}) as inputs for the benchmark generator. 
For each demand, we generate 500 samples and randomly downsample the human-annotated benchmark to match the number of generated samples for fair comparison.
Each experiment is repeated three times, and the average results are reported.
We use GPT-4o mini \citep{4o} as the default generator and also explore the performance of GPT-4o and Claude 3.5 Haiku \citep{claude}. The decoding temperature is set to 1.
To mitigate the self-enhancement bias \citep{llmasjudge} associated with LLM-as-a-judge, we substitute the generators with Qwen-Plus \citep{qwen} as the judge.

\subsection{Comparison with Human-annotated Benchmark}
As shown in Table~\ref{tab:main}, overall, \textsc{BenchMaker} achieves comparable performance to human-annotated benchmarks in terms of faithfulness and lexical\&semantic diversity. Meanwhile, \textsc{BenchMaker} outperforms them in all other metrics, especially in alignment, knowledge diversity, difficulty controllability and efficiency. 
The exceptional results achieved in these metrics comprehensively validate the reliability of the generated samples by \textsc{BenchMaker}.
\paragraph{Effectiveness.} 
The primary goal of benchmarking is to assign accurate scores to models under evaluation, facilitating capability differentiation. The benchmarking results of \textsc{BenchMaker} align closely with human-annotated benchmarks, with an average of 0.953 linear correlation (Pearson) and a remarkable 0.966 for rank-order correlation (Spearman), highlighting its outstanding effectiveness.
\paragraph{Robustness.}
Under evaluation demands where semantic equivalence is maintained but linguistic styles vary, the benchmarks exhibit nearly identical assessment efficacy, with an average Pearson correlation of 0.984. This demonstrates the robustness of \textsc{BenchMaker} to diverse inputs and ensures that users with different linguistic preferences can obtain consistent evaluation results.
\paragraph{Efficiency.}
The primary limitation of human-annotated benchmarks lies in their low construction efficiency. However, \textsc{BenchMaker} can generate a sample at an average cost of \$0.005 within 0.40 minutes. Furthermore, its efficiency is expected to continuously improve with the development of technology and hardware.

\paragraph{Generalizability.}
Experimental results demonstrate that \textsc{BenchMaker} exhibits strong generalization across different task types and generators. Notably, a more powerful model does not necessarily yield superior performance across all metrics. Compared to GPT-4o, GPT-4o mini proves to be a more cost-effective benchmark generator.

\subsection{Ablation Studies}
We validate the effectiveness of different techniques by sequentially integrating them to the Direct Prompt baseline on the MATH benchmark, as shown in Table~\ref{tab:main}.
\paragraph{Diversity.}
Compared to Direct Prompt, both AttrPrompt and In-batch Diversity Boosting effectively enhance lexical and semantic diversity. Noticeably, we observe that knowledge diversity remains unchanged, indicating that surface-level diversification does not necessarily equate to a broader assessment of knowledge and skills. Meanwhile, the diversity improvement leads to a slight drop in faithfulness, possibly because of the attributes constraints.
\paragraph{Faithfulness.} 
After applying Stepwise Self-correction and Conflict Guided Contrastive Discrimination, we observe a sustained and significant improvement in faithfulness. At the same time, we notice a reduction in the difficulty of the hardest subset, with the error rate decreasing from 0.659 to 0.557. We hypothesize that this may be due to the high error rate in labels when faithfulness is not ensured, which leads to an underestimation of model performance. Consequently, once the labels are corrected, the accuracy can better reflect the actual difficulty of the benchmark.
\paragraph{Difficulty Controllability.}
By treating the generator as the test-taker and using its error rate as the difficulty label, we achieve more precise control over sample difficulty (Spearman correlation of 0.403). Considering the previously observed weak difficulty perception of LLMs, we hypothesize that this improvement stems from the role shift, which requires the model to engage in explicit reasoning, along with the adoption of prediction-label inconsistency as an objective metric.

\begin{figure}[t]
\centering
\includegraphics[width=0.47\textwidth]{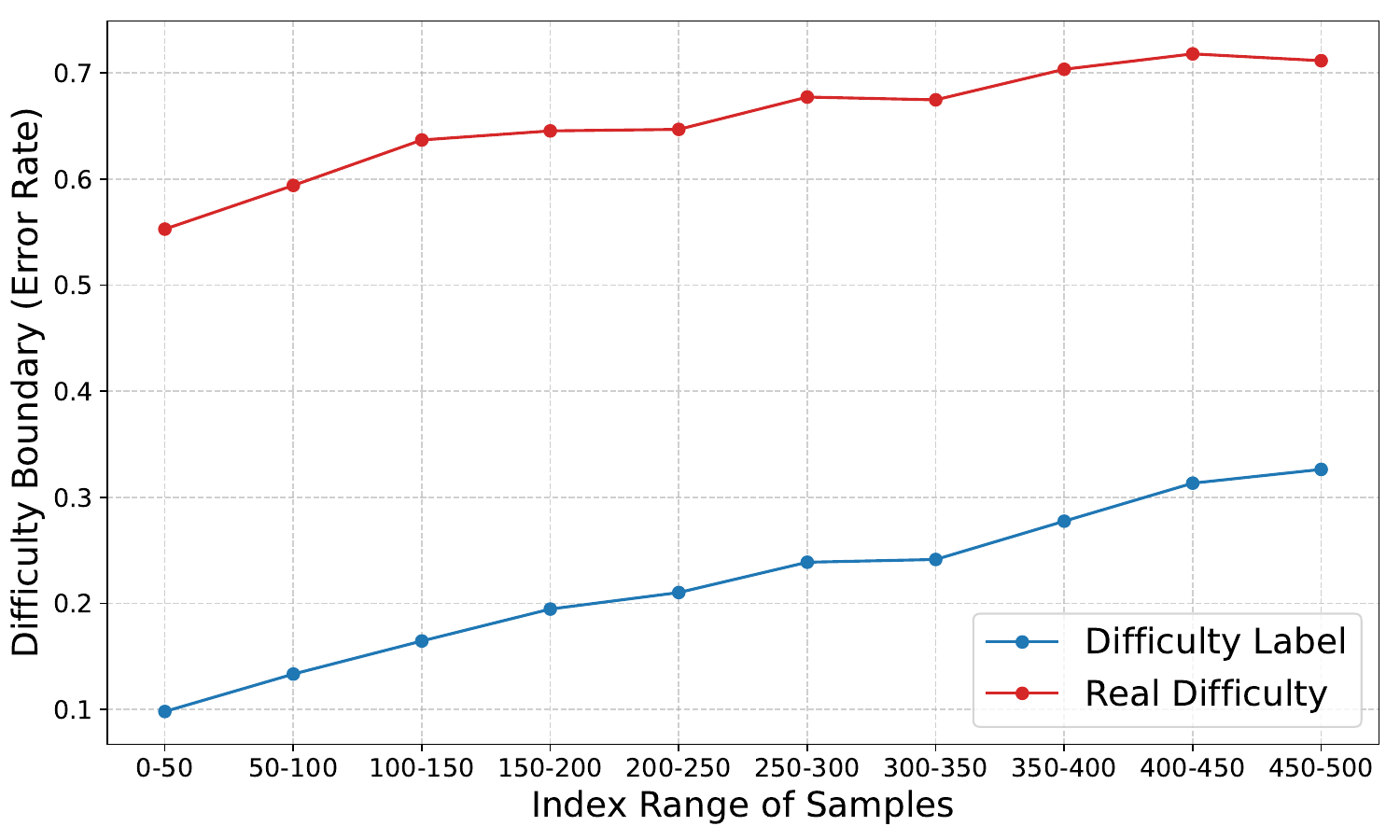}
\caption{Trends of real and labeled difficulty over the index.}
\label{fig:index}
\vspace{-5pt}
\end{figure}

\paragraph{Difficulty Boundary.}
With our proposed Difficulty Diffusion Mechanism and Difficulty Strategy Guidance, the difficulty boundary is significantly extended, as evidenced by an increase in error rate from 0.515 to 0.768, validating their effectiveness. Additionally, we analyze how the actual difficulty and difficulty labels evolve with the order of generated samples. As illustrated in Figure~\ref{fig:index}, both the difficulty label and actual difficulty exhibit a continuous upward trend. This not only confirms that Difficulty Diffusion Mechanism operates as intended but also visually demonstrates the strong consistency between difficulty label and actual difficulty.

\begin{figure}[!htb]
    \centering
    \subfigure[MMLU-Pro]{\includegraphics[width=0.48\hsize, height=0.27\hsize]{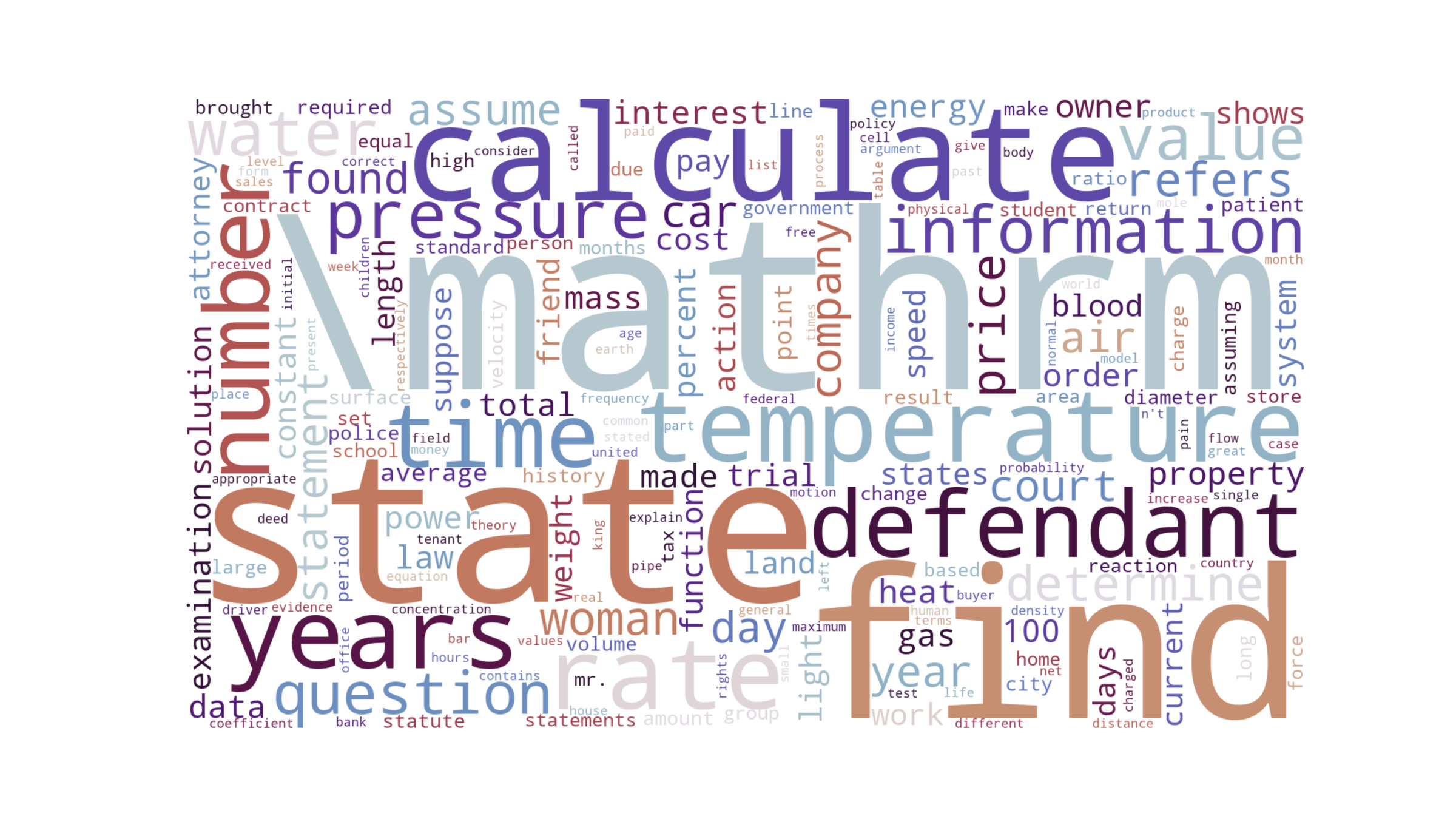}\label{fig: sub_figure1}} 
    \vspace{-3mm}
    \subfigure[\textsc{BenchMaker}]{\includegraphics[width=0.48\hsize, height=0.27\hsize]{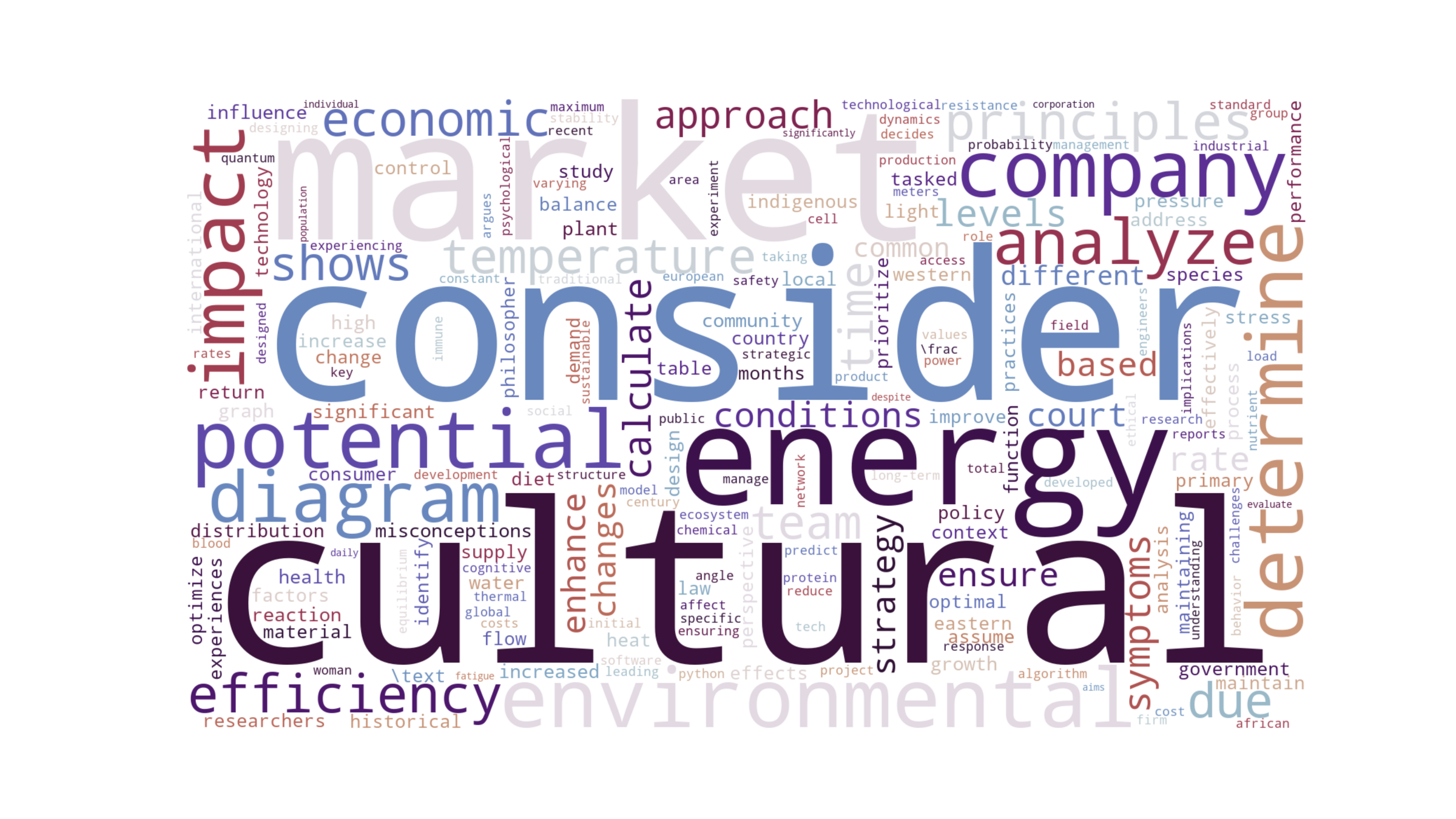}\label{fig: sub_figure2}} 
    \caption{Word cloud of MMLU-Pro and the benchmark generated by \textsc{BenchMaker} under similar assessment demands.}
    \label{fig:word}
\end{figure}

\begin{figure}[h]
\centering
\includegraphics[width=0.47\textwidth]{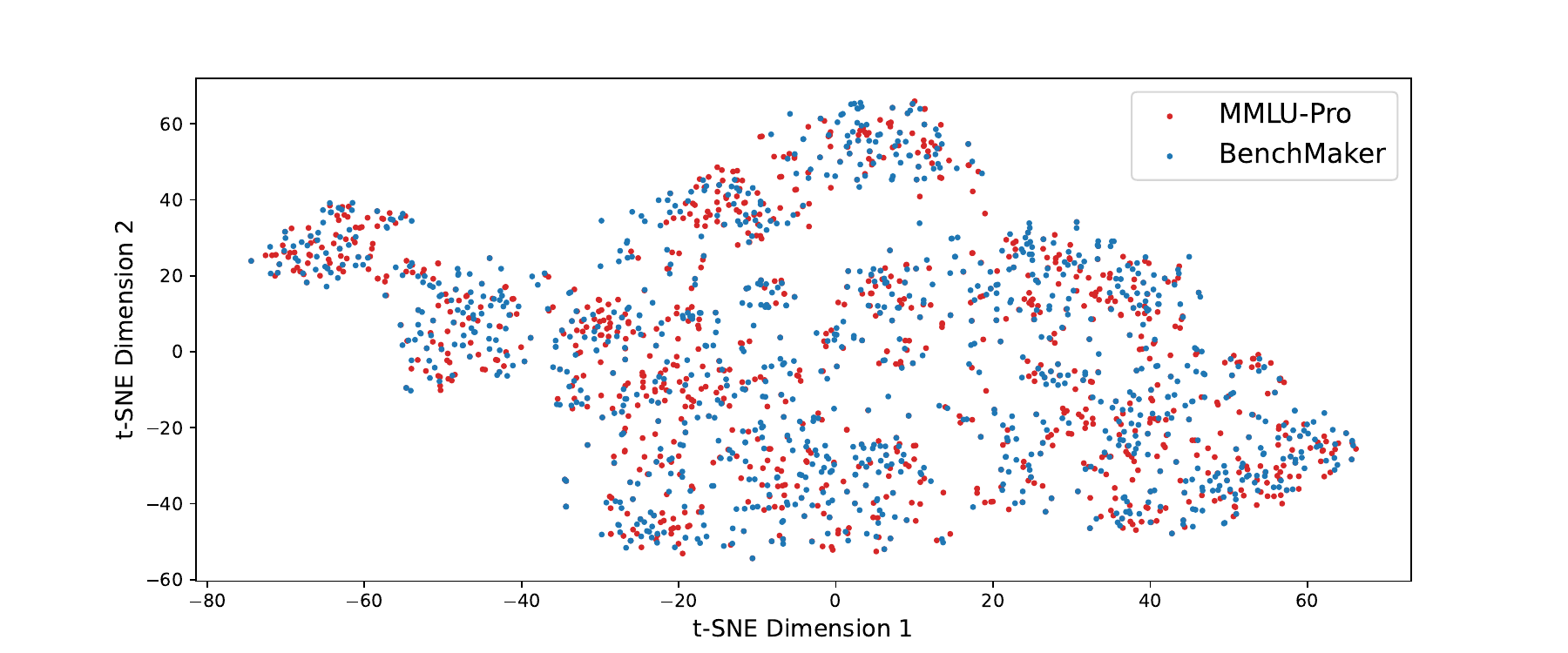}
\caption{T-SNE results on the text embeddings of benchmarks.}
\label{fig:tsne}
\end{figure}

\subsection{A Closer Look at the Generated Benchmark}
After metric analysis, we perform a more thorough examination of \textsc{BenchMaker}. 
Some of the generated samples are shown in Appendix~\ref{sec:example}.
\paragraph{Lexical and Semantic.}
First, despite the obvious differences in word distribution between the generated benchmark and MMLU-Pro (Figure~\ref{fig:word}), it remains closely aligned with the domains covered by MMLU-Pro, demonstrating strong task alignment.
Meanwhile, 
The semantic alignment between the two is more pronounced (Figure~\ref{fig:tsne}). Notably, the input demands (Appendix~\ref{sec:demand_mmlupro}) do not mention any information related to MMLU-Pro, effectively preventing the model from achieving a high degree of alignment by memorizing and replicating samples from MMLU-Pro.
\paragraph{Actual Error Rate.}
Although LLM-as-a-judge has provided an unbiased estimation of the benchmark’s faithfulness, we additionally conduct a manual check on 80 randomly selected samples. Our findings indicate that 3 samples have incorrect labels, 3 samples lack a correct candidate, resulting in an overall error rate of 7.5\%. Meanwhile, LLM-as-a-judge identifies 5 problematic samples, with 3 overlapping with human judgment. These results suggest that: (1) \textsc{BenchMaker} still has room for improvement in faithfulness; (2) LLM-as-a-judge can serve as a partial proxy for human evaluation.
\subsection{Reliability Estimation.}
Since faithfulness of the generated benchmark cannot be totally ensured, we are curious about the effects of incorrect samples:
Let $\Bar{a}$ and $\Bar{b}$ be the observed accuracies of two models $A$ and $B$ on the generated benchmark of size $N$, where a fraction $K$ of the samples are incorrect (which can be estimated by the LLM-as-a-judge). 
Suppose that $\bar{a}>\bar{b}$, we aim to estimate the probability that the observed ability rank is correct (the true accuracies satisfy $E[a] > E[b]$). See detailed derivation in Appendix~\ref{sec:prof}. Suppose $A$ and $B$ have the same accuracy $p$ on incorrect samples, we get:
\begin{equation}
E[a] = \frac{\bar{a} - K \cdot p}{1 - K}, \quad E[b] = \frac{\bar{b} - K \cdot p}{1 - K}
\end{equation}
and
\begin{equation}
E[a] - E[b] = \frac{\bar{a} - \bar{b}}{1 - K}
\end{equation}
Next, we perform hypothesis testing to assess the probability of \( E[a] > E[b] \). We assume that \( \bar{a} - \bar{b} \) follows a normal distribution. The \( z \)-score for the difference is:
\begin{equation}
\begin{split}
z =& \frac{(\bar{a} - \bar{b})/(1-K)}{\sqrt{(\bar{a}(1-\bar{a}) + \bar{b}(1-\bar{b}))/(N(1-K)^2)}} \\
=&\frac{(\bar{a} - \bar{b})\sqrt{N}}{\sqrt{\bar{a}(1-\bar{a}) + \bar{b}(1-\bar{b})}}
\end{split}
\end{equation}
where \( \Phi(z) \) is the cumulative distribution function of the standard normal distribution. 
The probability \( P(E[a] > E[b]) \) is given by the right-tail probability of the normal distribution:
\begin{equation}
P(E[a] > E[b]) = 1 - \Phi(z)
\label{eq:bound}
\end{equation}
where \( \Phi(z) \) is the cumulative distribution function of the standard normal distribution. 
We can assess the reliability of \textsc{BenchMaker} evaluation results using ~\eqref{eq:bound}.
Also, we notice that $K$ has the same scaling effect on both the numerator and denominator of the test statistic, thus does not alter the \( z \)-score. Consequently, as long as there is no bias, a certain proportion of noise in the benchmark will not affect the statistical significance of ability ranking.
\section*{Conclusions}
\label{sec:con}
The rapid advancement of large language models has driven an urgent demand for a generic benchmark generator. To this end, we first propose a comprehensive, automated, and unbiased evaluation framework to validate and optimize the reliability of benchmark generators.
Based on this, we develop the \textsc{BenchMaker} method for reliable, generic, and efficient benchmark generation.
Comprehensive experiments across multiple tasks and LLMs demonstrate that \textsc{BenchMaker} achieves human-aligned benchmark quality, with superior efficiency and generalization. 

\section*{Impact Statement}
This paper presents \textsc{BenchMaker}, an LLM-driven reliable , generic and efficient benchmark generator. There
are many potential societal consequences of our work, none
which we feel must be specifically highlighted here.
\paragraph{Ethics Statement.}
All of the datasets used in this study were publicly available. 
Multiple authors jointly conducted the manual check for the Actual Error Rate section, and no extra annotators were employed for our data collection.
We confirm that the datasets we used did not contain any harmful content and was consistent with their intended use (research). We have cited the datasets and relevant works used in this study.

\bibliography{acl_latex}

\begin{thebibliography}{43}
\expandafter\ifx\csname natexlab\endcsname\relax\def\natexlab#1{#1}\fi

\bibitem[{Anthropic()}]{claude}
Anthropic.
\newblock Claude 3.5.
\newblock \url{https://www.anthropic.com/news/3-5-models-and-computer-use}.

\bibitem[{Balloccu et~al.(2024)Balloccu, Schmidtov{\'{a}}, Lango, and Dusek}]{datacontamination}
Simone Balloccu, Patr{\'{\i}}cia Schmidtov{\'{a}}, Mateusz Lango, and Ondrej Dusek. 2024.
\newblock \href {https://aclanthology.org/2024.eacl-long.5} {Leak, cheat, repeat: Data contamination and evaluation malpractices in closed-source llms}.
\newblock In \emph{Proceedings of the 18th Conference of the European Chapter of the Association for Computational Linguistics, {EACL} 2024 - Volume 1: Long Papers, St. Julian's, Malta, March 17-22, 2024}, pages 67--93. Association for Computational Linguistics.

\bibitem[{Chan et~al.(2024)Chan, Wang, Yu, Mi, and Yu}]{persona}
Xin Chan, Xiaoyang Wang, Dian Yu, Haitao Mi, and Dong Yu. 2024.
\newblock \href {https://doi.org/10.48550/ARXIV.2406.20094} {Scaling synthetic data creation with 1,000,000,000 personas}.
\newblock \emph{CoRR}, abs/2406.20094.

\bibitem[{Chang et~al.(2024)Chang, Wang, Wang, Wu, Yang, Zhu, Chen, Yi, Wang, Wang, Ye, Zhang, Chang, Yu, Yang, and Xie}]{evalsurvey}
Yupeng Chang, Xu~Wang, Jindong Wang, Yuan Wu, Linyi Yang, Kaijie Zhu, Hao Chen, Xiaoyuan Yi, Cunxiang Wang, Yidong Wang, Wei Ye, Yue Zhang, Yi~Chang, Philip~S. Yu, Qiang Yang, and Xing Xie. 2024.
\newblock \href {https://doi.org/10.1145/3641289} {A survey on evaluation of large language models}.
\newblock \emph{{ACM} Trans. Intell. Syst. Technol.}, 15(3):39:1--39:45.

\bibitem[{Glazer et~al.(2024)Glazer, Erdil, Besiroglu, Chicharro, Chen, Gunning, Olsson, Denain, Ho, de~Oliveira~Santos, J{\"{a}}rviniemi, Barnett, Sandler, Vrzala, Sevilla, Ren, Pratt, Levine, Barkley, Stewart, Grechuk, Grechuk, Enugandla, and Wildon}]{saturation}
Elliot Glazer, Ege Erdil, Tamay Besiroglu, Diego Chicharro, Evan Chen, Alex Gunning, Caroline~Falkman Olsson, Jean{-}Stanislas Denain, Anson Ho, Emily de~Oliveira~Santos, Olli J{\"{a}}rviniemi, Matthew Barnett, Robert Sandler, Matej Vrzala, Jaime Sevilla, Qiuyu Ren, Elizabeth Pratt, Lionel Levine, Grant Barkley, Natalie Stewart, Bogdan Grechuk, Tetiana Grechuk, Shreepranav~Varma Enugandla, and Mark Wildon. 2024.
\newblock \href {https://doi.org/10.48550/ARXIV.2411.04872} {Frontiermath: {A} benchmark for evaluating advanced mathematical reasoning in {AI}}.
\newblock \emph{CoRR}, abs/2411.04872.

\bibitem[{Guo and Vosoughi(2023)}]{diverbias}
Xiaobo Guo and Soroush Vosoughi. 2023.
\newblock \href {https://doi.org/10.18653/V1/2023.EMNLP-MAIN.984} {Length does matter: Summary length can bias summarization metrics}.
\newblock In \emph{Proceedings of the 2023 Conference on Empirical Methods in Natural Language Processing, {EMNLP} 2023, Singapore, December 6-10, 2023}, pages 15869--15879. Association for Computational Linguistics.

\bibitem[{Hamming(1950)}]{hamming}
Richard~W Hamming. 1950.
\newblock Error detecting and error correcting codes.
\newblock \emph{The Bell system technical journal}, 29(2):147--160.

\bibitem[{Hendrycks et~al.(2021{\natexlab{a}})Hendrycks, Burns, Basart, Zou, Mazeika, Song, and Steinhardt}]{mmlu}
Dan Hendrycks, Collin Burns, Steven Basart, Andy Zou, Mantas Mazeika, Dawn Song, and Jacob Steinhardt. 2021{\natexlab{a}}.
\newblock \href {https://openreview.net/forum?id=d7KBjmI3GmQ} {Measuring massive multitask language understanding}.
\newblock In \emph{9th International Conference on Learning Representations, {ICLR} 2021, Virtual Event, Austria, May 3-7, 2021}. OpenReview.net.

\bibitem[{Hendrycks et~al.(2021{\natexlab{b}})Hendrycks, Burns, Kadavath, Arora, Basart, Tang, Song, and Steinhardt}]{math}
Dan Hendrycks, Collin Burns, Saurav Kadavath, Akul Arora, Steven Basart, Eric Tang, Dawn Song, and Jacob Steinhardt. 2021{\natexlab{b}}.
\newblock \href {https://datasets-benchmarks-proceedings.neurips.cc/paper/2021/hash/be83ab3ecd0db773eb2dc1b0a17836a1-Abstract-round2.html} {Measuring mathematical problem solving with the {MATH} dataset}.
\newblock In \emph{Proceedings of the Neural Information Processing Systems Track on Datasets and Benchmarks 1, NeurIPS Datasets and Benchmarks 2021, December 2021, virtual}.

\bibitem[{Huang et~al.(2024)Huang, Chen, Mishra, Zheng, Yu, Song, and Zhou}]{cannotselfcritic}
Jie Huang, Xinyun Chen, Swaroop Mishra, Huaixiu~Steven Zheng, Adams~Wei Yu, Xinying Song, and Denny Zhou. 2024.
\newblock \href {https://openreview.net/forum?id=IkmD3fKBPQ} {Large language models cannot self-correct reasoning yet}.
\newblock In \emph{The Twelfth International Conference on Learning Representations, {ICLR} 2024, Vienna, Austria, May 7-11, 2024}. OpenReview.net.

\bibitem[{Hurst et~al.(2024)Hurst, Lerer, Goucher, Perelman, Ramesh, Clark, Ostrow, Welihinda, Hayes, Radford et~al.}]{4o}
Aaron Hurst, Adam Lerer, Adam~P Goucher, Adam Perelman, Aditya Ramesh, Aidan Clark, AJ~Ostrow, Akila Welihinda, Alan Hayes, Alec Radford, et~al. 2024.
\newblock Gpt-4o system card.
\newblock \emph{arXiv preprint arXiv:2410.21276}.

\bibitem[{Ji et~al.(2023)Ji, Yu, Xu, Lee, Ishii, and Fung}]{SelfReflect}
Ziwei Ji, Tiezheng Yu, Yan Xu, Nayeon Lee, Etsuko Ishii, and Pascale Fung. 2023.
\newblock \href {https://doi.org/10.18653/V1/2023.FINDINGS-EMNLP.123} {Towards mitigating {LLM} hallucination via self reflection}.
\newblock In \emph{Findings of the Association for Computational Linguistics: {EMNLP} 2023, Singapore, December 6-10, 2023}, pages 1827--1843. Association for Computational Linguistics.

\bibitem[{Kaddour et~al.(2022)Kaddour, Lynch, Liu, Kusner, and Silva}]{caulearn}
Jean Kaddour, Aengus Lynch, Qi~Liu, Matt~J. Kusner, and Ricardo Silva. 2022.
\newblock \href {https://doi.org/10.48550/ARXIV.2206.15475} {Causal machine learning: {A} survey and open problems}.
\newblock \emph{CoRR}, abs/2206.15475.

\bibitem[{Lei et~al.(2023)Lei, Liu, Huang, He, Zhao, and Liu}]{s3eval}
Fangyu Lei, Qian Liu, Yiming Huang, Shizhu He, Jun Zhao, and Kang Liu. 2023.
\newblock \href {https://doi.org/10.48550/ARXIV.2310.15147} {S3eval: {A} synthetic, scalable, systematic evaluation suite for large language models}.
\newblock \emph{CoRR}, abs/2310.15147.

\bibitem[{Lewis et~al.(2020)Lewis, Perez, Piktus, Petroni, Karpukhin, Goyal, K{\"{u}}ttler, Lewis, Yih, Rockt{\"{a}}schel, Riedel, and Kiela}]{rag}
Patrick S.~H. Lewis, Ethan Perez, Aleksandra Piktus, Fabio Petroni, Vladimir Karpukhin, Naman Goyal, Heinrich K{\"{u}}ttler, Mike Lewis, Wen{-}tau Yih, Tim Rockt{\"{a}}schel, Sebastian Riedel, and Douwe Kiela. 2020.
\newblock \href {https://proceedings.neurips.cc/paper/2020/hash/6b493230205f780e1bc26945df7481e5-Abstract.html} {Retrieval-augmented generation for knowledge-intensive {NLP} tasks}.
\newblock In \emph{Advances in Neural Information Processing Systems 33: Annual Conference on Neural Information Processing Systems 2020, NeurIPS 2020, December 6-12, 2020, virtual}.

\bibitem[{Li et~al.(2024{\natexlab{a}})Li, Hu, Huang, Zhuang, Liu, Zhu, Shi, and Lin}]{perteval}
Jiatong Li, Renjun Hu, Kunzhe Huang, Yan Zhuang, Qi~Liu, Mengxiao Zhu, Xing Shi, and Wei Lin. 2024{\natexlab{a}}.
\newblock \href {https://doi.org/10.48550/ARXIV.2405.19740} {Perteval: Unveiling real knowledge capacity of llms with knowledge-invariant perturbations}.
\newblock \emph{CoRR}, abs/2405.19740.

\bibitem[{Li et~al.(2024{\natexlab{b}})Li, Huang, Zhuang, Shi, Cai, Xu, Wang, Zhang, Ke, and Cai}]{datasci}
Sihang Li, Jin Huang, Jiaxi Zhuang, Yaorui Shi, Xiaochen Cai, Mingjun Xu, Xiang Wang, Linfeng Zhang, Guolin Ke, and Hengxing Cai. 2024{\natexlab{b}}.
\newblock \href {https://doi.org/10.48550/ARXIV.2408.15545} {Scilitllm: How to adapt llms for scientific literature understanding}.
\newblock \emph{CoRR}, abs/2408.15545.

\bibitem[{Li et~al.(2024{\natexlab{c}})Li, Zhou, Yao, Li, Cao, Yang, Zhang, and Ma}]{solver}
Zenan Li, Zhi Zhou, Yuan Yao, Yu{-}Feng Li, Chun Cao, Fan Yang, Xian Zhang, and Xiaoxing Ma. 2024{\natexlab{c}}.
\newblock \href {https://doi.org/10.48550/ARXIV.2412.04857} {Neuro-symbolic data generation for math reasoning}.
\newblock \emph{CoRR}, abs/2412.04857.

\bibitem[{Liu et~al.(2023)Liu, Iter, Xu, Wang, Xu, and Zhu}]{geval}
Yang Liu, Dan Iter, Yichong Xu, Shuohang Wang, Ruochen Xu, and Chenguang Zhu. 2023.
\newblock \href {https://aclanthology.org/2023.emnlp-main.153} {G-eval: {NLG} evaluation using gpt-4 with better human alignment}.
\newblock In \emph{Proceedings of the 2023 Conference on Empirical Methods in Natural Language Processing, {EMNLP} 2023, Singapore, December 6-10, 2023}, pages 2511--2522. Association for Computational Linguistics.

\bibitem[{Long et~al.(2024)Long, Wang, Xiao, Zhao, Ding, Chen, and Wang}]{datasurvey2}
Lin Long, Rui Wang, Ruixuan Xiao, Junbo Zhao, Xiao Ding, Gang Chen, and Haobo Wang. 2024.
\newblock \href {https://doi.org/10.18653/V1/2024.FINDINGS-ACL.658} {On llms-driven synthetic data generation, curation, and evaluation: {A} survey}.
\newblock In \emph{Findings of the Association for Computational Linguistics, {ACL} 2024, Bangkok, Thailand and virtual meeting, August 11-16, 2024}, pages 11065--11082. Association for Computational Linguistics.

\bibitem[{Luo et~al.(2024{\natexlab{a}})Luo, Sun, Xu, Zhao, Lin, Lou, Chen, Tang, and Chen}]{flywheel1}
Haipeng Luo, Qingfeng Sun, Can Xu, Pu~Zhao, Qingwei Lin, Jianguang Lou, Shifeng Chen, Yansong Tang, and Weizhu Chen. 2024{\natexlab{a}}.
\newblock \href {https://doi.org/10.48550/ARXIV.2407.10627} {Arena learning: Build data flywheel for llms post-training via simulated chatbot arena}.
\newblock \emph{CoRR}, abs/2407.10627.

\bibitem[{Luo et~al.(2024{\natexlab{b}})Luo, Xu, Zhao, Sun, Geng, Hu, Tao, Ma, Lin, and Jiang}]{datacode}
Ziyang Luo, Can Xu, Pu~Zhao, Qingfeng Sun, Xiubo Geng, Wenxiang Hu, Chongyang Tao, Jing Ma, Qingwei Lin, and Daxin Jiang. 2024{\natexlab{b}}.
\newblock \href {https://openreview.net/forum?id=UnUwSIgK5W} {Wizardcoder: Empowering code large language models with evol-instruct}.
\newblock In \emph{The Twelfth International Conference on Learning Representations, {ICLR} 2024, Vienna, Austria, May 7-11, 2024}. OpenReview.net.

\bibitem[{Maheshwari et~al.(2024)Maheshwari, Ivanov, and Haddad}]{databench}
Gaurav Maheshwari, Dmitry Ivanov, and Kevin~El Haddad. 2024.
\newblock \href {https://doi.org/10.48550/ARXIV.2409.11968} {Efficacy of synthetic data as a benchmark}.
\newblock \emph{CoRR}, abs/2409.11968.

\bibitem[{Montahaei et~al.(2019)Montahaei, Alihosseini, and Baghshah}]{entropy}
Ehsan Montahaei, Danial Alihosseini, and Mahdieh~Soleymani Baghshah. 2019.
\newblock \href {http://arxiv.org/abs/1904.03971} {Jointly measuring diversity and quality in text generation models}.
\newblock \emph{CoRR}, abs/1904.03971.

\bibitem[{OpenAI()}]{text-embedding-ada-002}
OpenAI.
\newblock text-embedding-ada-002.
\newblock \url{https://platform.openai.com/docs/guides/embeddings}.

\bibitem[{Perez et~al.(2023)Perez, Ringer, Lukosiute, Nguyen, Chen, Heiner, Pettit, Olsson, Kundu, Kadavath, Jones, Chen, Mann, Israel, Seethor, McKinnon, Olah, Yan, Amodei, Amodei, Drain, Li, Tran{-}Johnson, Khundadze, Kernion, Landis, Kerr, Mueller, Hyun, Landau, Ndousse, Goldberg, Lovitt, Lucas, Sellitto, Zhang, Kingsland, Elhage, Joseph, Mercado, DasSarma, Rausch, Larson, McCandlish, Johnston, Kravec, Showk, Lanham, Telleen{-}Lawton, Brown, Henighan, Hume, Bai, Hatfield{-}Dodds, Clark, Bowman, Askell, Grosse, Hernandez, Ganguli, Hubinger, Schiefer, and Kaplan}]{modelwritten}
Ethan Perez, Sam Ringer, Kamile Lukosiute, Karina Nguyen, Edwin Chen, Scott Heiner, Craig Pettit, Catherine Olsson, Sandipan Kundu, Saurav Kadavath, Andy Jones, Anna Chen, Benjamin Mann, Brian Israel, Bryan Seethor, Cameron McKinnon, Christopher Olah, Da~Yan, Daniela Amodei, Dario Amodei, Dawn Drain, Dustin Li, Eli Tran{-}Johnson, Guro Khundadze, Jackson Kernion, James Landis, Jamie Kerr, Jared Mueller, Jeeyoon Hyun, Joshua Landau, Kamal Ndousse, Landon Goldberg, Liane Lovitt, Martin Lucas, Michael Sellitto, Miranda Zhang, Neerav Kingsland, Nelson Elhage, Nicholas Joseph, Noem{\'{\i}} Mercado, Nova DasSarma, Oliver Rausch, Robin Larson, Sam McCandlish, Scott Johnston, Shauna Kravec, Sheer~El Showk, Tamera Lanham, Timothy Telleen{-}Lawton, Tom Brown, Tom Henighan, Tristan Hume, Yuntao Bai, Zac Hatfield{-}Dodds, Jack Clark, Samuel~R. Bowman, Amanda Askell, Roger Grosse, Danny Hernandez, Deep Ganguli, Evan Hubinger, Nicholas Schiefer, and Jared Kaplan. 2023.
\newblock \href {https://doi.org/10.18653/V1/2023.FINDINGS-ACL.847} {Discovering language model behaviors with model-written evaluations}.
\newblock In \emph{Findings of the Association for Computational Linguistics: {ACL} 2023, Toronto, Canada, July 9-14, 2023}, pages 13387--13434. Association for Computational Linguistics.

\bibitem[{Tam et~al.(2024)Tam, Wu, Tsai, Lin, Lee, and Chen}]{format}
Zhi~Rui Tam, Cheng{-}Kuang Wu, Yi{-}Lin Tsai, Chieh{-}Yen Lin, Hung{-}yi Lee, and Yun{-}Nung Chen. 2024.
\newblock \href {https://aclanthology.org/2024.emnlp-industry.91} {Let me speak freely? {A} study on the impact of format restrictions on large language model performance}.
\newblock In \emph{Proceedings of the 2024 Conference on Empirical Methods in Natural Language Processing: {EMNLP} 2024 - Industry Track, Miami, Florida, USA, November 12-16, 2024}, pages 1218--1236. Association for Computational Linguistics.

\bibitem[{Tao et~al.(2024)Tao, Lin, Chen, Li, Wu, Li, Jin, Huang, Tao, and Zhou}]{flywheel2}
Zhengwei Tao, Ting{-}En Lin, Xiancai Chen, Hangyu Li, Yuchuan Wu, Yongbin Li, Zhi Jin, Fei Huang, Dacheng Tao, and Jingren Zhou. 2024.
\newblock \href {https://doi.org/10.48550/ARXIV.2404.14387} {A survey on self-evolution of large language models}.
\newblock \emph{CoRR}, abs/2404.14387.

\bibitem[{Thakur et~al.(2024)Thakur, Choudhary, Ramayapally, Vaidyanathan, and Hupkes}]{judgingjudge}
Aman~Singh Thakur, Kartik Choudhary, Venkat~Srinik Ramayapally, Sankaran Vaidyanathan, and Dieuwke Hupkes. 2024.
\newblock \href {https://doi.org/10.48550/ARXIV.2406.12624} {Judging the judges: Evaluating alignment and vulnerabilities in llms-as-judges}.
\newblock \emph{CoRR}, abs/2406.12624.

\bibitem[{Vallat(2018)}]{pengouin}
Raphael Vallat. 2018.
\newblock \href {https://doi.org/10.21105/JOSS.01026} {Pingouin: statistics in python}.
\newblock \emph{J. Open Source Softw.}, 3(31):1026.

\bibitem[{Wang et~al.(2024{\natexlab{a}})Wang, Zhu, Ren, Liu, Li, Zhang, Zhang, Wu, Zhan, Liu, and Wang}]{datasurvey1}
Ke~Wang, Jiahui Zhu, Minjie Ren, Zeming Liu, Shiwei Li, Zongye Zhang, Chenkai Zhang, Xiaoyu Wu, Qiqi Zhan, Qingjie Liu, and Yunhong Wang. 2024{\natexlab{a}}.
\newblock \href {https://doi.org/10.48550/ARXIV.2410.12896} {A survey on data synthesis and augmentation for large language models}.
\newblock \emph{CoRR}, abs/2410.12896.

\bibitem[{Wang et~al.(2023{\natexlab{a}})Wang, Wei, Schuurmans, Le, Chi, Narang, Chowdhery, and Zhou}]{sc}
Xuezhi Wang, Jason Wei, Dale Schuurmans, Quoc~V. Le, Ed~H. Chi, Sharan Narang, Aakanksha Chowdhery, and Denny Zhou. 2023{\natexlab{a}}.
\newblock \href {https://openreview.net/forum?id=1PL1NIMMrw} {Self-consistency improves chain of thought reasoning in language models}.
\newblock In \emph{The Eleventh International Conference on Learning Representations, {ICLR} 2023, Kigali, Rwanda, May 1-5, 2023}. OpenReview.net.

\bibitem[{Wang et~al.(2023{\natexlab{b}})Wang, Kordi, Mishra, Liu, Smith, Khashabi, and Hajishirzi}]{SelfIns}
Yizhong Wang, Yeganeh Kordi, Swaroop Mishra, Alisa Liu, Noah~A. Smith, Daniel Khashabi, and Hannaneh Hajishirzi. 2023{\natexlab{b}}.
\newblock \href {https://doi.org/10.18653/V1/2023.ACL-LONG.754} {Self-instruct: Aligning language models with self-generated instructions}.
\newblock In \emph{Proceedings of the 61st Annual Meeting of the Association for Computational Linguistics (Volume 1: Long Papers), {ACL} 2023, Toronto, Canada, July 9-14, 2023}, pages 13484--13508. Association for Computational Linguistics.

\bibitem[{Wang et~al.(2024{\natexlab{b}})Wang, Ma, Zhang, Ni, Chandra, Guo, Ren, Arulraj, He, Jiang, Li, Ku, Wang, Zhuang, Fan, Yue, and Chen}]{mmlupro}
Yubo Wang, Xueguang Ma, Ge~Zhang, Yuansheng Ni, Abhranil Chandra, Shiguang Guo, Weiming Ren, Aaran Arulraj, Xuan He, Ziyan Jiang, Tianle Li, Max Ku, Kai Wang, Alex Zhuang, Rongqi Fan, Xiang Yue, and Wenhu Chen. 2024{\natexlab{b}}.
\newblock \href {https://doi.org/10.48550/ARXIV.2406.01574} {Mmlu-pro: {A} more robust and challenging multi-task language understanding benchmark}.
\newblock \emph{CoRR}, abs/2406.01574.

\bibitem[{Wu et~al.(2024)Wu, Huang, Gao, Chen, Zhang, Wan, Zhou, Zhang, Gao, Xiao, and Sun}]{unigen}
Siyuan Wu, Yue Huang, Chujie Gao, Dongping Chen, Qihui Zhang, Yao Wan, Tianyi Zhou, Xiangliang Zhang, Jianfeng Gao, Chaowei Xiao, and Lichao Sun. 2024.
\newblock \href {https://doi.org/10.48550/ARXIV.2406.18966} {Unigen: {A} unified framework for textual dataset generation using large language models}.
\newblock \emph{CoRR}, abs/2406.18966.

\bibitem[{Yang et~al.(2024)Yang, Yang, Hui, Zheng, Yu, Zhou, Li, Li, Liu, Huang, Dong, Wei, Lin, Tang, Wang, Yang, Tu, Zhang, Ma, Yang, Xu, Zhou, Bai, He, Lin, Dang, Lu, Chen, Yang, Li, Xue, Ni, Zhang, Wang, Peng, Men, Gao, Lin, Wang, Bai, Tan, Zhu, Li, Liu, Ge, Deng, Zhou, Ren, Zhang, Wei, Ren, Liu, Fan, Yao, Zhang, Wan, Chu, Liu, Cui, Zhang, Guo, and Fan}]{qwen}
An~Yang, Baosong Yang, Binyuan Hui, Bo~Zheng, Bowen Yu, Chang Zhou, Chengpeng Li, Chengyuan Li, Dayiheng Liu, Fei Huang, Guanting Dong, Haoran Wei, Huan Lin, Jialong Tang, Jialin Wang, Jian Yang, Jianhong Tu, Jianwei Zhang, Jianxin Ma, Jianxin Yang, Jin Xu, Jingren Zhou, Jinze Bai, Jinzheng He, Junyang Lin, Kai Dang, Keming Lu, Keqin Chen, Kexin Yang, Mei Li, Mingfeng Xue, Na~Ni, Pei Zhang, Peng Wang, Ru~Peng, Rui Men, Ruize Gao, Runji Lin, Shijie Wang, Shuai Bai, Sinan Tan, Tianhang Zhu, Tianhao Li, Tianyu Liu, Wenbin Ge, Xiaodong Deng, Xiaohuan Zhou, Xingzhang Ren, Xinyu Zhang, Xipin Wei, Xuancheng Ren, Xuejing Liu, Yang Fan, Yang Yao, Yichang Zhang, Yu~Wan, Yunfei Chu, Yuqiong Liu, Zeyu Cui, Zhenru Zhang, Zhifang Guo, and Zhihao Fan. 2024.
\newblock \href {https://doi.org/10.48550/ARXIV.2407.10671} {Qwen2 technical report}.
\newblock \emph{CoRR}, abs/2407.10671.

\bibitem[{Yu et~al.(2024)Yu, Jiang, Shi, Yu, Liu, Zhang, Kwok, Li, Weller, and Liu}]{datamath}
Longhui Yu, Weisen Jiang, Han Shi, Jincheng Yu, Zhengying Liu, Yu~Zhang, James~T. Kwok, Zhenguo Li, Adrian Weller, and Weiyang Liu. 2024.
\newblock \href {https://openreview.net/forum?id=N8N0hgNDRt} {Metamath: Bootstrap your own mathematical questions for large language models}.
\newblock In \emph{The Twelfth International Conference on Learning Representations, {ICLR} 2024, Vienna, Austria, May 7-11, 2024}. OpenReview.net.

\bibitem[{Yu et~al.(2023)Yu, Zhuang, Zhang, Meng, Ratner, Krishna, Shen, and Zhang}]{attr}
Yue Yu, Yuchen Zhuang, Jieyu Zhang, Yu~Meng, Alexander~J. Ratner, Ranjay Krishna, Jiaming Shen, and Chao Zhang. 2023.
\newblock \href {http://papers.nips.cc/paper\_files/paper/2023/hash/ae9500c4f5607caf2eff033c67daa9d7-Abstract-Datasets\_and\_Benchmarks.html} {Large language model as attributed training data generator: {A} tale of diversity and bias}.
\newblock In \emph{Advances in Neural Information Processing Systems 36: Annual Conference on Neural Information Processing Systems 2023, NeurIPS 2023, New Orleans, LA, USA, December 10 - 16, 2023}.

\bibitem[{Zellers et~al.(2019)Zellers, Holtzman, Bisk, Farhadi, and Choi}]{hellaswag}
Rowan Zellers, Ari Holtzman, Yonatan Bisk, Ali Farhadi, and Yejin Choi. 2019.
\newblock \href {https://doi.org/10.18653/V1/P19-1472} {Hellaswag: Can a machine really finish your sentence?}
\newblock In \emph{Proceedings of the 57th Conference of the Association for Computational Linguistics, {ACL} 2019, Florence, Italy, July 28- August 2, 2019, Volume 1: Long Papers}, pages 4791--4800. Association for Computational Linguistics.

\bibitem[{Zheng et~al.(2023)Zheng, Chiang, Sheng, Zhuang, Wu, Zhuang, Lin, Li, Li, Xing, Zhang, Gonzalez, and Stoica}]{llmasjudge}
Lianmin Zheng, Wei{-}Lin Chiang, Ying Sheng, Siyuan Zhuang, Zhanghao Wu, Yonghao Zhuang, Zi~Lin, Zhuohan Li, Dacheng Li, Eric~P. Xing, Hao Zhang, Joseph~E. Gonzalez, and Ion Stoica. 2023.
\newblock \href {http://papers.nips.cc/paper\_files/paper/2023/hash/91f18a1287b398d378ef22505bf41832-Abstract-Datasets\_and\_Benchmarks.html} {Judging llm-as-a-judge with mt-bench and chatbot arena}.
\newblock In \emph{Advances in Neural Information Processing Systems 36: Annual Conference on Neural Information Processing Systems 2023, NeurIPS 2023, New Orleans, LA, USA, December 10 - 16, 2023}.

\bibitem[{Zhu et~al.(2024{\natexlab{a}})Zhu, Chen, Wang, Gong, Yang, and Xie}]{dyval}
Kaijie Zhu, Jiaao Chen, Jindong Wang, Neil~Zhenqiang Gong, Diyi Yang, and Xing Xie. 2024{\natexlab{a}}.
\newblock \href {https://openreview.net/forum?id=gjfOL9z5Xr} {Dyval: Dynamic evaluation of large language models for reasoning tasks}.
\newblock In \emph{The Twelfth International Conference on Learning Representations, {ICLR} 2024, Vienna, Austria, May 7-11, 2024}. OpenReview.net.

\bibitem[{Zhu et~al.(2024{\natexlab{b}})Zhu, Wang, Zhao, Xu, and Xie}]{dyval2}
Kaijie Zhu, Jindong Wang, Qinlin Zhao, Ruochen Xu, and Xing Xie. 2024{\natexlab{b}}.
\newblock \href {https://doi.org/10.48550/ARXIV.2402.14865} {Dyval 2: Dynamic evaluation of large language models by meta probing agents}.
\newblock \emph{CoRR}, abs/2402.14865.

\bibitem[{Zhu et~al.(2018)Zhu, Lu, Zheng, Guo, Zhang, Wang, and Yu}]{selfbleu}
Yaoming Zhu, Sidi Lu, Lei Zheng, Jiaxian Guo, Weinan Zhang, Jun Wang, and Yong Yu. 2018.
\newblock \href {https://doi.org/10.1145/3209978.3210080} {Texygen: {A} benchmarking platform for text generation models}.
\newblock In \emph{The 41st International {ACM} {SIGIR} Conference on Research {\&} Development in Information Retrieval, {SIGIR} 2018, Ann Arbor, MI, USA, July 08-12, 2018}, pages 1097--1100. {ACM}.

\end{thebibliography}
\clearpage
\onecolumn
\appendix
\section{Derivation of the Probability for \( E[a] > E[b] \)}
\label{sec:prof}
In this appendix, we present a detailed derivation of the probability that the expected accuracy \( E[a] \) of model A is greater than the expected accuracy \( E[b] \) of model B in a noiseless benchmark, given the noisy benchmark observations. We will use the following notation throughout:

\begin{itemize}
    \item \( N \): The total number of samples in the benchmark.
    \item \( K \): The proportion of samples with incorrect labels, i.e., the noise ratio, where \( K \in [0,1] \).
    \item \( \bar{a} \): The observed accuracy of model A on the noisy benchmark.
    \item \( \bar{b} \): The observed accuracy of model B on the noisy benchmark.
    \item \( p \): The probability that both models predict the incorrect label correctly on a noisy sample. This probability is assumed to be identical for both models on incorrect labels.
    \item \( E[a] \): The expected accuracy of model A on the noiseless benchmark.
    \item \( E[b] \): The expected accuracy of model B on the noiseless benchmark.
\end{itemize}

Given these notations, our goal is to determine the probability that \( E[a] > E[b] \) based on the noisy observed accuracies \( \bar{a} \) and \( \bar{b} \).

\subsection*{Step 1: Relating \( \bar{a} \) and \( \bar{b} \) to \( E[a] \) and \( E[b] \)}

From the given setup, we know that the observed accuracies \( \bar{a} \) and \( \bar{b} \) can be written as a weighted average of the expected accuracies \( E[a] \) and \( E[b] \) on the correct samples, and the probability \( p \) on the incorrect samples. Specifically, the formulas for \( \bar{a} \) and \( \bar{b} \) are:

\[
\bar{a} = (1 - K) \cdot E[a] + K \cdot p
\]
\[
\bar{b} = (1 - K) \cdot E[b] + K \cdot p
\]

These equations express the observed accuracy of each model as the weighted average of the correct label samples and the noisy (incorrect label) samples. The weight \( (1 - K) \) represents the proportion of correct labels, and \( K \) represents the proportion of incorrect labels.

\subsection*{Step 2: Solving for \( E[a] \) and \( E[b] \)}

To isolate \( E[a] \) and \( E[b] \), we rearrange the above equations:

\[
E[a] = \frac{\bar{a} - K \cdot p}{1 - K}
\]
\[
E[b] = \frac{\bar{b} - K \cdot p}{1 - K}
\]

Thus, \( E[a] \) and \( E[b] \) are directly related to the observed accuracies \( \bar{a} \) and \( \bar{b} \), and the noise ratio \( K \).

\subsection*{Step 3: Comparing \( E[a] \) and \( E[b] \)}

To determine the probability that \( E[a] > E[b] \), we first compute the difference between \( E[a] \) and \( E[b] \):

\[
E[a] - E[b] = \frac{\bar{a} - K \cdot p}{1 - K} - \frac{\bar{b} - K \cdot p}{1 - K}
\]

Simplifying this expression:

\[
E[a] - E[b] = \frac{\bar{a} - \bar{b}}{1 - K}
\]

Thus, \( E[a] > E[b] \) if and only if \( \bar{a} - \bar{b} > 0 \), which indicates that the observed accuracy of model A must be greater than that of model B in the noisy benchmark for \( E[a] \) to exceed \( E[b] \) in the noiseless benchmark.

\subsection*{Step 4: Statistical Hypothesis Testing}

To quantify the probability of \( E[a] > E[b] \), we perform a hypothesis test on \( \bar{a} - \bar{b} \), assuming that both \( \bar{a} \) and \( \bar{b} \) are derived from binomial distributions (since they represent the correct classification probabilities on the noisy benchmark). We assume that the difference \( \bar{a} - \bar{b} \) follows a normal distribution under certain conditions (via the Central Limit Theorem). Thus, the expected value of \( \bar{a} - \bar{b} \) is:

\[
\mathbb{E}[\bar{a} - \bar{b}] = \frac{\bar{a} - \bar{b}}{1 - K}
\]

The variance of \( \bar{a} - \bar{b} \), assuming independent samples, is given by:

\begin{equation}
\begin{split}
\text{Var}[\bar{a} - \bar{b}] =& \frac{\text{Var}[\bar{a}] + \text{Var}[\bar{b}]}{(1 - K)^2} \\
=&\frac{\bar{a}(1-\bar{a})/N +\bar{b}(1-\bar{b})/N}{(1 - K)^2}
\end{split}
\end{equation}

where \( \text{Var}[\bar{a}] \) and \( \text{Var}[\bar{b}] \) are the variances of the observed accuracies of models A and B, respectively. These variances can be computed from the binomial distributions underlying \( \bar{a} \) and \( \bar{b} \).

Now, we define the \( z \)-score for the observed difference \( \bar{a} - \bar{b} \) as follows:

\begin{equation}
\begin{split}
z =& \frac{\mathbb{E}[\bar{a} - \bar{b}]}{\sqrt{\text{Var}[\bar{a} - \bar{b}]}}\\
=&\frac{(\bar{a} - \bar{b})/(1-K)}{\sqrt{(\bar{a}(1-\bar{a}) + \bar{b}(1-\bar{b}))/(N(1-K)^2)}} \\
=&\frac{(\bar{a} - \bar{b})\sqrt{N}}{\sqrt{\bar{a}(1-\bar{a}) + \bar{b}(1-\bar{b})}}
\end{split}
\end{equation}

This \( z \)-score follows a standard normal distribution. The probability that \( E[a] > E[b] \) is the probability that \( \bar{a} - \bar{b} > 0 \), which is equivalent to:

\[
P(\bar{a} - \bar{b} > 0) = P(z > 0) = 1 - \Phi(z)
\]

where \( \Phi(z) \) is the cumulative distribution function (CDF) of the standard normal distribution. Thus, the \( p \)-value is given by:

\[
p\text{-value} = 1 - \Phi(z)
\]

\subsection*{Step 5: Conclusion}

To summarize, the probability that \( E[a] > E[b] \) in the noiseless benchmark, given the noisy benchmark observations, is determined by the observed accuracy difference \( \bar{a} - \bar{b} \) and the noise ratio \( K \). The probability is computed using a hypothesis test on \( \bar{a} - \bar{b} \), assuming it follows a normal distribution. The final formula for this probability is:

\[
P(E[a] > E[b]) = 1 - \Phi(z)
\]

where \( \Phi(z) \) is the CDF of the standard normal distribution and \( z \) is the computed \( z \)-score. This result allows us to assess the likelihood that model A has a higher expected accuracy than model B in the noiseless benchmark based on noisy observations.

\section{Unsuccessful Attempts for Optimizing Benchmark Generator}
\label{sec:unsuccess}
\subsection{Faithfulness}
We explored the widely studied self-correction strategy to improve the faithfulness of benchmarks. Specifically, for each generated sample, the model first acts as a judge and then refines samples it deems insufficiently faithful. However, our preliminary results indicate that while this approach yields minor improvements in mathematical tasks, it provides little benefit for tasks such as MMLU-Pro and instead introduces additional computational overhead.

\subsection{Difficulty Controllability}
As previously mentioned, we attempted to have the model generate samples with specified difficulty levels, but the resulting samples exhibited low difficulty differentiation. To address this, we further explored having the model assess the difficulty of its generated samples. However, this strategy yielded promising results only on the MATH task.

\subsection{Difficulty Diffusion Mechanism}
Previous studies \citep{mmlupro} have attempted to increase question difficulty by expanding the number of answer choices. However, our experiments show that scaling up the number of candidates quickly reaches a saturation point. We hypothesize that this is due to the model’s difficulty in generating a large number of sufficiently deceptive distractors.

\subsection{Diversity}
To enhance sample diversity, in addition to AttrPrompt, we experimented with assigning different personas \citep{persona} to the model and instructing it to generate characteristic samples based on its assigned persona. However, we found that this approach was not particularly effective for the MATH task, especially in semantic diversity.

\section{Data from Huggingface}
\label{sec:huggingface}
We obtained information on open-source model releases and download counts from the Hugging Face API (\texttt{from huggingface\_hub import HfApi}). Since the number of open-source model releases far exceeds that of closed-source models, we use the former to represent the "Number of Language Model Releases." Additionally, as Hugging Face does not provide monthly download counts for each model, we use the historical total downloads of models released within a given statistical period as the total downloads for that period. The corresponding code is shown below.

\section{Difficulty Levels}
\label{sec:dif_level}
\begin{itemize}
    \item \textbf{Level 1}: The simplest, equivalent to lower-grade elementary school
\item \textbf{Level 2}: Relatively simple, equivalent to upper-grade elementary school
\item \textbf{Level 3}: Simple, equivalent to middle school
\item \textbf{Level 4}: Average, equivalent to high school
\item \textbf{Level 5}: Slightly difficult, equivalent to university student
\item \textbf{Level 6}: Difficult, equivalent to Master's
\item \textbf{Level 7}: Quite difficult, equivalent to PhD student
\item \textbf{Level 8}: Very difficult, equivalent to professor
\item \textbf{Level 9}: Extremely difficult, equivalent to field expert
\item \textbf{Level 10}: Most difficult, equivalent to top human level or beyond human level
\end{itemize}

\section{Benchmarking Model List}
\label{sec:bench_model_list}
\begin{itemize}
\item \textbf{phoenix-inst-chat-7b}: \url{https://huggingface.co/FreedomIntelligence/phoenix-inst-chat-7b}
\item \textbf{vicuna-7b-v1.3}: \url{https://huggingface.co/lmsys/vicuna-7b-v1.3}
\item \textbf{Qwen2.5-3B}: \url{https://huggingface.co/Qwen/Qwen2.5-3B}
\item \textbf{phi-2}: \url{https://huggingface.co/microsoft/phi-2}
\item \textbf{Phi-3.5-mini-instruct}: \url{https://huggingface.co/microsoft/Phi-3.5-mini-instruct}
\item \textbf{Yi-1.5-6B-Chat}: \url{https://huggingface.co/01-ai/Yi-1.5-6B-Chat}
\item \textbf{Qwen2.5-7B}: \url{https://huggingface.co/Qwen/Qwen2.5-7B}
\item \textbf{vicuna-7b-v1.5}: \url{https://huggingface.co/lmsys/vicuna-7b-v1.5}
\item \textbf{Qwen2-1.5B-Instruct}: \url{https://huggingface.co/Qwen/Qwen2-1.5B-Instruct}
\item \textbf{phoenix-inst-chat-7b-v1.1}: \url{https://huggingface.co/FreedomIntelligence/phoenix-inst-chat-7b-v1.1}
\item \textbf{Qwen-Plus}: \url{https://huggingface.co/Qwen} 
\item \textbf{GPT-3.5 turbo}: \url{https://openai.com/index/gpt-3-5-turbo-fine-tuning-and-api-updates/}
\end{itemize}

\section{Details of Difficulty Diffusion Mechanism}
\label{sec:ddd}
Given that the LLM has a certain level of difficulty perception, we iteratively select the more challenging samples according to $\beta$ from the generated ones as difficulty references, and instruct the LLM to generate a more difficult sample. 
Specifically,
To prevent reference samples from becoming overly fixed, which may lead to homogenization in generated samples, we adopt the following strategy:
\begin{enumerate}
    \item We track the number of times each sample \( x_i \) has been used as a reference sample, denoted as \( t_i \), and compute a calibrated difficulty label:
    \begin{equation}
        \text{Calibrate\_Difficulty} = \text{Difficulty\_Label} \times 0.9^{t_i / \text{Reference\_Number}}
    \end{equation}
    The samples are then sorted based on this adjusted difficulty.

    \item Each time, we select \( 2 \times \text{Reference\_Number} \) samples with the highest \( \text{Calibrate\_Difficulty} \) as candidates. From this pool, we randomly sample \( \text{Reference\_Number} \) as reference samples and shuffle their order.
\end{enumerate}
Our preliminary experiments indicate a positive correlation between problem difficulty and \( \text{Reference\_Number} \). In our experiments, we set \( \text{Reference\_Number} \) to 8.
This allows the sample difficulty to rise continuously through diffusion.

\begin{figure*}[h]
\centering
\includegraphics[width=0.95\textwidth]{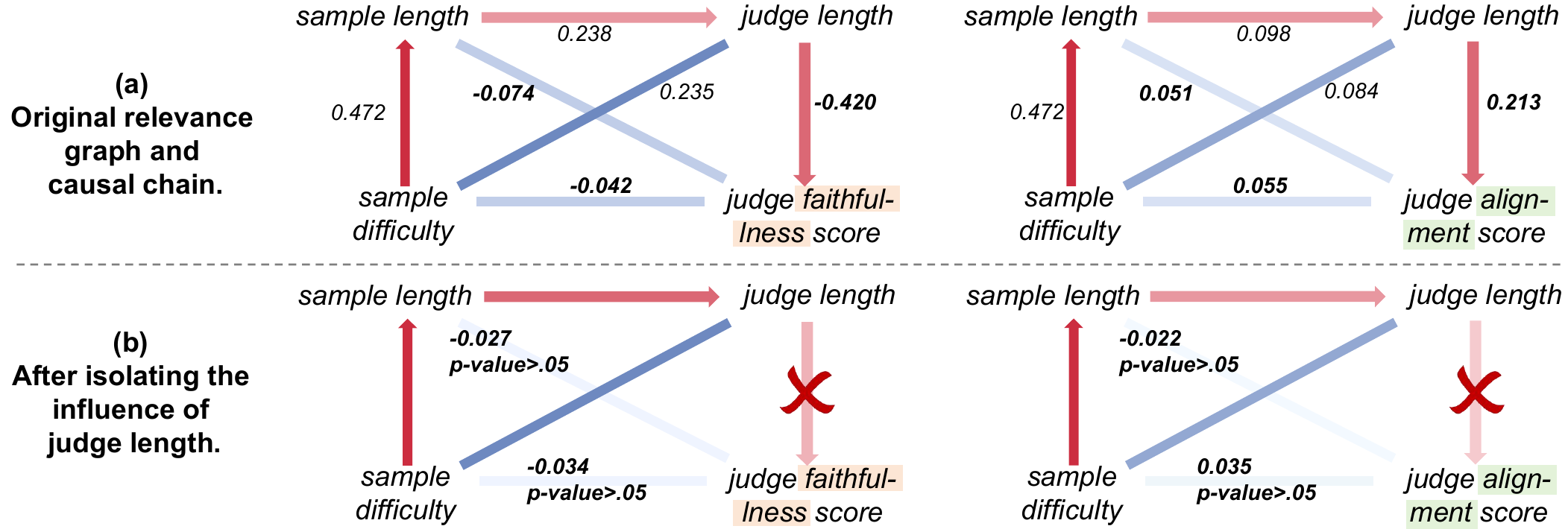}
\vspace{-10pt}
\caption{{Pearson correlations among key factors of benchmark evaluation and LLM (GPT-4o mini) judge scores (faithfulness and alignment). The most relevant path of each subject is highlighted in red to show the possible causal chain.}}
\vspace{-7pt}
\label{fig:chain_4omini}
\end{figure*}

\section{Converting Benchmark Sample Format}
\paragraph{MCQ to OTG Format.}
\label{sec:convert}
By removing the options from the samples and using the solution and answer corresponding to the correct option as the ground truth, we can easily transform the MCQ-style benchmark generated based on MATH assessment demands into an open-ended text generation (OTG) benchmark. Comparing these two benchmark formats, we find that the OTG format makes the questions more challenging (error rate: $0.865 > 0.768$) and results in lower knowledge diversity (hamming distance: $0.365 < 0.403$). We attribute this to the model's inability to rely on option cues to answer certain questions, which leads to a large portion of the knowledge vector being zero, thereby reducing knowledge diversity. Additionally, we observe a decline in the benchmark’s effectiveness (pearson: $0.915 < 0.935$), which we hypothesize is indirectly caused by the drop in knowledge diversity.

\paragraph{OTG to MCQ Format.}
\label{sec:MATH_convert}
To convert the MATH benchmark into a MCQ format, we employed GPT-4o, GPT-4o mini, GPT-3.5 turbo, and Claude-3.5 Haiku to answer MATH problems, sampling 10 responses per model with a temperature of 1. We identified the three most frequently occurring incorrect answers as distractors while retaining the correct answer's rationale. If the number of incorrect answers was insufficient, we supplemented them using GPT-4o mini. Through this process, we successfully transformed the MATH benchmark from an OTG format to an MCQ format.

\section{Examples of the Generated Samples}
\label{sec:example}
\textbf{MATH:}
\begin{lstlisting}
Example 1:
A researcher is studying the distribution of three specific proteins in a cell. 
There are 4 locations within the cell where each protein can be present. 
However, due to experimental conditions, at least one protein must be present 
in each location. In how many different ways can the proteins be distributed 
in the cell, considering overlap in presence is allowed?
A. 2187
B. 2401
C. 4096
D. 2048

Label:B


Example 2:
Find the smallest positive integer \\( n \\) such that \\( n \\) is divisible
by 6, 10, and 15, and \\( n \\equiv 2 \\pmod{4} \\).
A. 120
B. 20
C. 90
D. 30

Label:D
\end{lstlisting}

\textbf{MMLU-Pro:}
\begin{lstlisting}
Example 1:
A 45-year-old woman with type 2 diabetes decides to improve her health by
adopting a low-carbohydrate, high-protein diet, starting a daily
30-minute brisk walk routine, and taking a new medication that
increases insulin sensitivity. She also begins consuming a herbal
supplement believed to enhance energy levels. After two months, she
notices an increase in fatigue, frequent headaches, and unexplained
weight gain. What is the most likely reason for her symptoms?
A. Low-carbohydrate diet leading to nutritional deficiencies
B. Brisk walk routine causing excessive physical exertion
C. Medication side effects causing insulin fluctuations
D. Herbal supplement causing hormonal imbalance
E. Increased protein intake causing kidney strain
F. Inadequate hydration from dietary changes
G. Overconsumption of high-protein foods leading to weight gain
H. Lack of fiber intake affecting metabolism
I. Decrease in carbohydrate intake causing energy depletion
J. Stress from lifestyle changes impacting health

Label:C


Example 2:
An architect is designing a complex apartment building, which features a 
series of irregularly shaped balconies. The layout of one of the building's
wings is depicted in the accompanying diagram. Each balcony's area is defined
by the function \( f(x) = 3x^2 + 2x + 1 \) over the interval [0, 2] meters,
representing a horizontal cross-section. The total length of the wing is
10 meters, and each balcony occurs at every meter along this length, 
aligned perpendicularly. To meet safety regulations, the 
architect needs to ensure that the probability of a randomly selected 
balcony having an area greater than 8 square meters is at least 0.5. 
Calculate the probability that a randomly selected balcony from this
wing has an area greater than 8 square meters, using integration
to determine the areas and probabilities involved. Consider potential
pitfalls like incorrect integral setup or probability interpretation.
A. 0.1
B. 0.2
C. 0.3
D. 0.4
E. 0.5
F. 0.6
G. 0.7
H. 0.8
I. 0.9
J. 1.0

Label:J
\end{lstlisting}

\textbf{HellaSwag:}
\begin{lstlisting}
During a family reunion, Mark is honored with the 'Outstanding Contributor'
award for his recent volunteer efforts in the community. As he stands
in front of his relatives, he expresses heartfelt gratitude towards everyone
who supported him but fails to mention his younger sister, Lily, who 
organized the charity event that helped him earn this recognition. 
After the ceremony, Lily watches Mark celebrate with others, her face
a mix of pride and disappointment. When Mark approaches her, excitedly
asking, 'Did you see me win? I couldn't have done it without your help!'
Given Lily's conflicting feelings about being overlooked, how is she
most likely to respond?
A. 'Congratulations, Mark! I'm really proud of you! But I can't help 
feeling a bit overshadowed since I organized the event.'  
B. 'Wow, Mark! You totally deserve this! Yet, it's tough for me to 
celebrate when my efforts went unnoticed.'  
C. 'That was an amazing award, Mark! I'm happy for you! However, it 
stings that my contribution was overlooked.'  
D. 'I'm so thrilled for you, Mark! Your achievement is incredible! 
But it feels a little unfair that I didn't get a shoutout.'

Label:C
\end{lstlisting}

\section{Prompt List}
\textbf{LLM as Faithfulness Judge:}
\label{sec:prompt_judge}
\begin{lstlisting}
You are an expert who excels at analyzing whether a given response 
correctly answers a provided question.

**Question:**
{{question}}

**Response to be Checked:**
{{response}}

Please note that the given question may be unsolvable, have a unique solution, multiple solutions,
etc. Therefore, you should carefully analyze the correctness of
the response to be checked based on the given question.

Here are the rules to strictly follow when analyzing the
correctness of a response:
1. **Step-by-Step Analysis**: Analyze the response step by step, reviewing
the reasoning and correctness of each step. For every step, first **restate
and summarize** the reasoning logic and conclusion presented in the response,
then analyze the correctness of that specific step.
2. **Focus on Evaluation**: Remember that your primary mission is to determine
whether the reasoning process is correct. Avoid attempting to solve the
problem yourself. Instead, focus strictly on analyzing the correctness
of the response's reasoning process, one step at a time.
3. **Avoid Premature Judgments**: Do not rush to make judgments (such as
claiming the response is flawed or completely correct) at the beginning.
Ensure your evaluation is based on a thorough step-by-step analysis before
arriving at a conclusion.
4. **Reverse Validation**: After completing the step-by-step analysis,
substitute the answer back into the original problem and perform
reverse validation of the parameters to cross-verify the correctness
of the response.
After completing your analysis, please provide your judgment on the correctness
of the response, as well as your confidence level in that judgment.

Your output should follow the template and example below:
Analyses:{Your detailed analyses}
Judgement:{0: You think both the final answer of the response is wrong;
0.5: You think the reasoning path has some mistakes, but the final
answer of the response is correct; 1: You agree with the reasoning path
and the final answer of the response}

##Example##
Analyses:{Your detailed analyses}
Judgement:1
##Example End##

Now begin with "Analyses:"
\end{lstlisting}

\textbf{LLM as Comparison-based Faithfulness Judge:}
\label{sec:prompt_cmp_judge}
\begin{lstlisting}
You are a knowledgeable expert with the task of analyzing the quality
of a given question and its candidate answers.

###Question
{{question}}

###Candidate 1:
{{can1}}

###Candidate 2:
{{can2}}

###Your task: Correctness Analysis
1. Analyze whether the question is correct, reasonable, and clearly stated.
2. For the given question, analyze whether the provided ###Candidate 1 and
###Candidate 2 are correct step by step sequentially.
(Do not favor a candidate just because it is long; evaluate candidates strictly
based on correctness.)
3. Based on the above analysis, output your judgment of the question
quality according to the following scale:
    0 point indicate an incorrect question with ambiguities and no uniquely 
    suitable answer among the options.
    0.5 point indicates a minor error in the question, but there is 
    still a uniquely suitable answer among the options.
    1 point indicate no errors in the question, with one uniquely correct
    answer among the options.
4. Please also output your chosen correct option
You should follow the template
below to output:
"##Faithfulness:{{score}}##, ##Label:{{}}##" (e.g., ##Faithfulness:2##,##Label:B##).
Please note that if you believe there is no correct option or there are multiple
correct options, output ##Faithfulness:0##, ##Label:None##.

You should begin your response with "Correctness Analysis".

\end{lstlisting}

\textbf{LLM as Relevance Judge:}
\label{sec:prompt_relevance_judge}
\begin{lstlisting}
You are an expert who excels at analyzing whether a given question can be used to assess a specific ability.

**Question:**
{{question}}

**Ability:**
{{ability}}

You should first carefully analyze what abilities the given question can be used to test.
Based on this analysis, compare it with the given abilities.
After completing your analysis, please provide your judgment on whether the given question can be used to test the given ability, as well as your confidence in that judgment.

Your output should follow the template below:
Analyses:{Your detailed analyses}
Judgement:
{output 0 if: You believe the given question is completely unable to test the given ability;
output 0.5 if: You believe the given question is primarily meant to test other abilities, but can also test the given ability to some extent;
output 1 if: You believe the given question primarily tests the given ability.}

Now begin with "Analyses:"
\end{lstlisting}

\paragraph{Directly Prompting LLM as Generic Benchmark Generator:}
\label{sec:prompt_direct}
Notably, before allowing the LLM to formally generate the benchmark, we first require it to produce descriptions for each part of the sample based on the assessment demands, including Task Description, Query Description, and Option Description. This helps the model better understand and align with the assessment demands, ensuring higher-quality and more consistent benchmark generation.
\begin{lstlisting}
You are a knowledgeable benchmark creator.
Your task is to generate a creative questions based on the provided Task Description, Query Description, Option Description, Generation Guidelines, and Output Description to help build a benchmark that assesses the given task.

### Task Description:
{{task define}}

### Query Description:
{{query define}}

### Option Description:
{{option define}}

### Generation Guidelines:
1. Analyze the given task and think step-by-step about the content needed to construct the question, begin with "Analyses:".
2. Generate the question content, begin with "Question:".
3. Generate 10 candidates, with only one as the right option. Begin with "Candidates:".
4. Generate the index of the right option, begin with "Right Option:".

### Output Description:
Strictly follow the template below to generate your sample.
**Template**
##Analyses:## {{You analyze the provided attributes and outline the process for constructing the question to be generated.}}
##Question:## {{Your generated question content}}
##Candidates:##
{{Your generated Candidates}}
##Right Option:##{{Index of the right option, e.g., B}}
**Template End**

Attention: You need to **strictly follow the template** and don't generate any other contents. Begin your response with "##Analyses:## "
\end{lstlisting}

\section{Examples of the Generated Difficulty Strategies}
\label{sec:strategies}
\textbf{MATH:}
\begin{lstlisting}
Strategy 1:
Complexity of Biological Concept is Basic
Complexity of Biological Concept is Intermediate
Complexity of Biological Concept is Advanced

Strategy 2:
Required Reasoning Steps set as Single-step
Required Reasoning Steps set as Multi-step (2-3 steps)
Required Reasoning Steps set as Multi-step (4-6 steps)
Required Reasoning Steps set as More than 6 steps

Strategy 3:
Familiarity with the Topic is Common
Familiarity with the Topic is Uncommon
Familiarity with the Topic is Rare

Strategy 4:
Type of Biological Data Analysis is Qualitative
Type of Biological Data Analysis is Quantitative
Type of Biological Data Analysis is Advanced Data Interpretation

Strategy 5:
Application of Concepts is Direct
Application of Concepts is Modified
Application of Concepts is Novel

Strategy 6:
Integration Across Biological Disciplines is Single-discipline
Integration Across Biological Disciplines is Cross-disciplinary
Integration Across Biological Disciplines is Interdisciplinary

Strategy 7:
Depth of Required Knowledge is Surface-level
Depth of Required Knowledge is In-depth
Depth of Required Knowledge is Comprehensive
\end{lstlisting}

\textbf{Prompt of \textsc{BenchMaker}:}
\label{sec:prompt_benchmaker}
\begin{lstlisting}
You are a knowledgeable benchmark creator.
Your task is to generate a creative question based on the provided Task Description, Query Description, Option Description, General Attributes Descriptions, Difficulty Strategies Description, Generation Guidelines, and Output Description to help build a benchmark that assesses the given task.

### Overall Task Description:
{{original task}}


### Detailed Task Description:
{{task define}}


### Query Description:
{{query define}}


### Option Description:
{{option define}}


### General Attributes Description:
You can refer to the following attributes and their corresponding values to construct questions, which means the questions you generate should ideally align with some of these attributes.
Please note, if you find any conflicting or confusing parts among the attributes listed, you may disregard them.
{{attribute define}}


### Difficulty Strategies Description:
Your generated questions should meet the following difficulty attribute requirements. If you find conflicts among these requirements, you may choose to selectively ignore them.
{{difficulty attribute define}}


### Difficulty Description:
The following are some samples (0 or several).
Please ensure that the difficulty level of the samples you generate is harder than these examples.
The samples you generate should aim to assess different knowledge and skills compared to the given samples.
The format of given samples are not what you should follow.
**Please ensure that the sample you create differ substantially from the following samples, so as to maintain diversity in the resulting benchmark.**
{{demonstrations}}


### Generation Guidelines:
**Stage 1: Analyze**
In this stage, you should analyze following the steps below and begin with "##Analyses:##". **You need to clearly articulate the analysis content for each step**, which means after completing Stage 1, you should have already produced a question that meets the requirements along with a correct and unique answer.
1-1. Analyze the general attributes, difficulty attributes and difficulty description, and think step-by-step about the content needed to construct the question. **Please use your imagination and avoid any obvious overlap with the given samples, either in the specific knowledge points being tested or in the format.**
1-2. Start by drafting your question. If you discover any issues with the question or any overlapping parts between the generated question and the given samples during this process, feel free to revise it.
1-3. Think through what the correct answer should be. If you discover any issues during this process, repeat the entire Stage 1 process from the beginning.
1-4. Identify the plausible and potentially misleading incorrect options that could serve as distractors (at least nine). If you discover any issues during this process, repeat the entire Stage 1 process from the beginning.
1-5. Reevaluate your proposed question, answer and options to ensure that: the question meet the given attributes and Difficulty Description (you should compare the generated samples and given samples to verify this); the answer is both correct and unique. If it does not meet these criteria or you are not sure about this, repeat the entire Stage 1 process from the beginning.

**Stage 2: Generate Sample**
In this stage, you should give your generated sample in the right template based on the analyses above.
2-1. Generate the question content, begin with "##Question:##".
2-2. Generate a step-by-step reasoning process and the corresponding correct answer. Begin with "##Reasoning Path:##". If you find an issue with the question, return to Step 2-1 to regenerate the question.
2-3. Generate {{OptionNum}} candidates, with only one as the right option. Begin with "##Candidates:##".
2-4. Generate the index of the right option, begin with "##Right Option:##".

### Output Description:
Strictly follow the template below to generate your sample.
**Template**
##Analyses:## {{You analyze the provided attributes and outline the process for constructing the question to be generated.}}
##Question:## {{Your generated question content}}
##Reasoning Path:## {{Your step-by-step reasoning process}}
##Candidates:##
{{CandidatesDemo}}
##Right Option:##{{Index of the right option, e.g., B}}
**Template End**


Attention: You need to **strictly follow the template** and don't generate any other contents. Begin your response with "##Analyses:##\n1-1. "
\end{lstlisting}

\section{Assessment Demands List}
\label{sec:demands}
\textbf{MATH:}
\label{sec:demand_math}
\begin{lstlisting}
Subset Name: Prealgebra
Assessment Demands:Prealgebra

Subset Name: Algebra
Assessment Demands:Algebra

Subset Name: Number Theory
Assessment Demands:Number Theory

Subset Name: Counting & Probability
Assessment Demands:Counting & Probability

Subset Name: Geometry
Assessment Demands:Geometry

Subset Name: Intermediate Algebra
Assessment Demands:Intermediate Algebra

Subset Name: Precalculus
Assessment Demands:Precalculus
\end{lstlisting}
\textbf{MMLU-Pro:}
\label{sec:demand_mmlupro}
\begin{lstlisting}
Subset Name: psychology
Assessment Demands:This benchmark is designed to assess **psychology** abilities while simultaneously evaluating knowledge understanding and complex reasoning skills, using **ten multiple-choice questions** as the evaluation format

Subset Name: philosophy
Assessment Demands:This benchmark is designed to assess **philosophy** abilities while simultaneously evaluating knowledge understanding and complex reasoning skills, using **ten multiple-choice questions** as the evaluation format

Subset Name: health
Assessment Demands:This benchmark is designed to assess **health** abilities while simultaneously evaluating knowledge understanding and complex reasoning skills, using **ten multiple-choice questions** as the evaluation format

Subset Name: history
Assessment Demands:This benchmark is designed to assess **history** abilities while simultaneously evaluating knowledge understanding and complex reasoning skills, using **ten multiple-choice questions** as the evaluation format

Subset Name: business
Assessment Demands:This benchmark is designed to assess **business** abilities while simultaneously evaluating knowledge understanding and complex reasoning skills, using **ten multiple-choice questions** as the evaluation format

Subset Name: physics
Assessment Demands:This benchmark is designed to assess **physics** abilities while simultaneously evaluating knowledge understanding and complex reasoning skills, using **ten multiple-choice questions** as the evaluation format

Subset Name: engineering
Assessment Demands:This benchmark is designed to assess **engineering** abilities while simultaneously evaluating knowledge understanding and complex reasoning skills, using **ten multiple-choice questions** as the evaluation format

Subset Name: chemistry
Assessment Demands:This benchmark is designed to assess **chemistry** abilities while simultaneously evaluating knowledge understanding and complex reasoning skills, using **ten multiple-choice questions** as the evaluation format

Subset Name: math
Assessment Demands:This benchmark is designed to assess **math** abilities while simultaneously evaluating knowledge understanding and complex reasoning skills, using **ten multiple-choice questions** as the evaluation format

Subset Name: computer science
Assessment Demands:This benchmark is designed to assess **computer science** abilities while simultaneously evaluating knowledge understanding and complex reasoning skills, using **ten multiple-choice questions** as the evaluation format

Subset Name: biology
Assessment Demands:This benchmark is designed to assess **biology** abilities while simultaneously evaluating knowledge understanding and complex reasoning skills, using **ten multiple-choice questions** as the evaluation format

Subset Name: economics
Assessment Demands:This benchmark is designed to assess **economics** abilities while simultaneously evaluating knowledge understanding and complex reasoning skills, using **ten multiple-choice questions** as the evaluation format

Subset Name: law
Assessment Demands:This benchmark is designed to assess **law** abilities while simultaneously evaluating knowledge understanding and complex reasoning skills, using **ten multiple-choice questions** as the evaluation format
\end{lstlisting}

\textbf{HellaSwag:}
\label{sec:demand_hella}
\begin{lstlisting}
Subset Name: NLI
Assessment Demands:The task is to evaluate the model's commonsense natural language inference ability. Specifically, each question should present a concrete scenario, and the model should select the most likely event from the options based on a series of inferences.
\end{lstlisting}

\end{document}